\documentclass[a4paper,fleqn]{cas-dc}

\usepackage[numbers,sort&compress]{natbib}
\usepackage{hyperref}
\usepackage{algorithm}
\usepackage{algpseudocode}

\def\tsc#1{\csdef{#1}{\textsc{\lowercase{#1}}\xspace}}
\tsc{WGM}
\tsc{QE}
\tsc{EP}
\tsc{PMS}
\tsc{BEC}
\tsc{DE}

\begin{document}
	\let\WriteBookmarks\relax
	\def\floatpagepagefraction{1}
	\def\textpagefraction{.001}
	\shorttitle{Modalities Should Talk to Each Other}
	\shortauthors{S.M.H Hashemi et~al.}
	
	\title [mode = title]{Modalities Should Talk to Each Other: Dual-Stream Multimodal Learning for Long-Horizon Influenza Forecasting}

	\author[1]{Seyed Mohammad Hossein Hashemi}[type=editor,
	auid=000,bioid=1,
	]
	\fnmark[1]
	
	\credit{Conceptualization of this study, Methodology, Investigation, Software, Visualization, Writing Original Draft}

	\author[1]{Mohsen Hooshmand}[type=editor,
	]
	
	\cormark[1]
	\fnmark[1]

	\credit{Conceptualization of this study, Methodology, Supervision, Validation, Resources, Writing Review \& Editing}

	\author[1]{Parvin Razzaghi}[type=editor,
	]
	
	\fnmark[1]
	
	\credit{Conceptualization of this study, Methodology, Supervision, Validation, Writing Review \& Editing}
	
	\affiliation[1]{organization={Department of Computer Science \& Information Technology, Institute for Advanced Studies in Basic Sciences (IASBS)},
		city={Zanjan},
		country={Iran}}

	\nonumnote{* Corresponding author.
	}
	\nonumnote{\hspace{6pt}E-mail addresses: mh.hashemi@iasbs.ac.ir (S.M.H. Hashemi), mohsen.hooshmand@iasbs.ac.ir (M. Hooshmand), p.razzaghi@iasbs.ac.ir (P. Razzaghi)}

	\begin{abstract}
		Forecasting long-range influenza-like illness (ILI) matters for public health readiness. Publicly available surveillance datasets typically pair numeric epidemiological signals with textual information that is noisy, loosely structured, only indirectly related to near-term trends, and often lagged relative to the numeric signal. Fusing the two therefore requires careful design. We propose Dual-Stream Attention (DSA), a multimodal deep learning framework that forecasts 12-week-ahead ILI activity from a 36-week multimodal history by letting the numerical and textual streams condition each other. Using the Time-MMD health-domain dataset, DSA separately encodes the two modalities with a Transformer-based numerical encoder and a domain-adapted headline encoder, then couples them through a bidirectional Cross-Modal Attention (CMA) mechanism: the text (news headlines) conditions the interpretation of the numeric signal and vice versa. The CMA output then passes to a causal temporal model for forecasting. Evaluated across ten random seeds, DSA achieves a median test MSE of 0.416, versus 0.668, 0.607, and 0.851 for iTransformer, TaTS, and GPT4MTS -- mean-error reductions of 54.95\%, 37.29\%, and 67.23\%, with paired Cohen's $d$ of 0.555, 0.337, and 0.345 -- and ranks first in 100\% of bootstrap draws. It also has substantially lower worst-window error than all baselines. On an external-geography dataset, DSA again ranks first among nine evaluated baselines. Ablations show the advantage does not depend on text-encoder choice or language-model fine-tuning, and that bidirectional attention outperforms either direction alone. Finally, perturbation-based faithfulness analysis shows the learned CMA is functionally informative under targeted masking, with a stronger effect in the text-to-numerical direction.
	\end{abstract}


	\maketitle
	
	\section{Introduction}

	Influenza-like illness (ILI) forecasting is an important component of infectious-disease surveillance because reliable estimates of near- and medium-term activity can support preparedness, resource allocation, and the interpretation of rapidly changing seasonal dynamics. The forecasting problem is difficult for two complementary reasons. First, ILI is a strongly seasonal and non-stationary process whose future trajectory depends on the recent epidemiological state\cite{shaman2012forecasting}. Second, the information available to surveillance systems is not restricted to numerical measurements: weekly surveillance environments are accompanied by textual signals describing symptoms, circulating viruses, public-health observations, and other contextual information\cite{brownstein2009digital}. A forecasting system that uses only the numerical series therefore discards a potentially informative part of the evidence available at the forecast origin.
	
	Recent advances in time-series forecasting have produced strong numerical models based on linear decompositions and Transformer architectures, including DLinear\cite{zeng2023transformers}, PatchTST\cite{nie2022time}, and iTransformer\cite{liu2024itransformer}. In parallel, multimodal and language-model-based approaches have begun to incorporate textual information alongside structured time series. This creates a specific methodological gap for surveillance forecasting: the two modalities are available at the same forecast origin, but their information content need not be equally reliable, equally relevant, or expressed on the same temporal scale. A weekly narrative may describe an emerging epidemiological change before it is fully visible in the numerical surveillance series, while numerical trends may also provide the context needed to interpret whether a textual signal is epidemiologically meaningful. Consequently, treating the modalities as fixed features and combining them only after independent encoding does not explicitly model how the state of one modality should influence the interpretation of the other.
	
	We address this gap with \emph{Dual-Stream Attention} (DSA), a multimodal forecasting framework designed to make this interaction explicit. DSA maintains separate numerical and text streams, represents the numerical history with a Transformer encoder and the weekly textual context with BioLinkBERT, and couples the streams through a bidirectional Cross-Modal Attention (CMA) module. The two attention directions provide complementary conditioning pathways: the numerical stream queries the textual stream, while the textual stream queries the numerical stream. The resulting representation is subsequently processed by a causal temporal Transformer and directly decoded into a 12-week forecast. Thus, rather than assuming that the two modalities should contribute equally or independently, DSA lets their representations be conditioned on one another before temporal aggregation and forecasting.
	
	The evaluation is designed to distinguish an architectural effect from a favorable random initialization or a narrow choice of baseline. We use the Time-MMD Health\_US benchmark with a 36-week lookback and a 12-week forecasting horizon, and compare DSA with strong unimodal time-series forecasters, text-augmented fusion models, and an LLM-based forecasting model. A two-stage protocol first screens the broader candidate set and then evaluates the strongest representatives across ten independent random seeds. Beyond average forecasting error, we examine tail-risk, behavior, epidemic-phase robustness, model ranking under bootstrap resampling, and external validation on Time-MMD Health\_Africa.
	
	The results show that DSA is consistently strong across these complementary evaluations. On Health\_US, its median test MSE is 0.4163 and it is included in the Model Confidence Set in the principal epidemic regimes. The paired comparisons favor DSA in all ten seeds against iTransformer, TaTS, and GPT4MTS, with Cohen's $d$ values of 0.555, 0.337, and 0.345, respectively. Importantly, the advantage is not limited to central tendency: DSA has markedly lower tail-risk than the competing models and remains the best model in the preliminary external Health\_Africa evaluation. The phase-stratified analysis clarifies where this advantage is concentrated: DSA achieves its clearest margin over every baseline in the peak phase, which is both the most operationally consequential regime and, plausibly, the one in which the textual signal is most informative, while also being the hardest phase to forecast for all four models. In the remaining regimes — off-season, rising, and declining — DSA is not always the single best model, but it stays on par with whichever baseline leads and remains among the top performers rather than falling behind. This distinction matters because a deployed forecasting system does not know in advance which epidemiological phase the next window belongs to, so a model that is consistently competitive across phases, and ahead of every baseline on average, offers more practical value than one that wins decisively in a single regime. We therefore interpret DSA as a robust overall architecture rather than as a model that dominates under every epidemiological condition.
	
	This study makes four main contributions:
	\begin{enumerate}
		\item We introduce DSA, a multimodal ILI forecasting architecture built on the intuition that noisy, temporally offset surveillance text should inform the numerical trend without being trusted unconditionally alongside it: DSA performs explicit bidirectional cross-modal conditioning between numerical epidemiological history and concurrent weekly text through CMA, allowing each modality's representation to be reweighted in light of the other rather than fused by unconditional concatenation.
		\item We provide a multi-seed evaluation against strong unimodal, multimodal, and LLM-based baselines, complemented by tail-risk, horizon-wise, phase-stratified, bootstrap rank-probability, and Model Confidence Set analyses.
		\item We investigate the source of DSA's advantage through controlled ablations of the fusion mechanism, its bidirectional design, language-model fine-tuning, and text encoder choice, and test preliminary generalization to a second geography.
		\item We evaluate the functional faithfulness of the learned cross-modal attention through targeted perturbations, showing that highly attended historical information produces greater forecasting degradation than random or weakly attended information, particularly in the Text$\rightarrow$Numerical direction, while acknowledging a recency confound that limits causal interpretation.
	\end{enumerate}
	
	The remainder of the paper reviews related forecasting and multimodal approaches, describes the proposed architecture and experimental protocol, presents the main results and robustness analyses, and concludes with the limitations and implications of the findings.

	\section{Related work}

	Work relevant to DSA is organized into four categories reflecting the progression from unimodal numerical forecasting to multimodal epidemiological prediction: numerical time-series forecasting provides the unimodal backbone; external-information and text-augmented forecasting establish how auxiliary information can complement numerical dynamics; LLM-based forecasting represents a newer approach to incorporating language into time-series models; and ILI/epidemic forecasting provides the disease-specific context. This organization distinguishes established approaches to numerical and epidemiological forecasting from the comparatively less explored problem of learning interactions between structured epidemic signals and semantic health text.
	
	\subsection{Unimodal time-series forecasting}
	Time-series forecasting has recently been dominated by Transformer-based architectures\cite{wu2021autoformer, zhou2022fedformer, wu2022timesnet, kitaev2020reformer, zhang2023crossformer, liu2022non, nie2022time, liu2024itransformer}. PatchTST\cite{nie2022time} treats sub-series patches as tokens, reducing sequence length and letting the model attend across coarser temporal units. iTransformer \cite{liu2024itransformer} inverts the usual Transformer treatment by embedding each variate, rather than each timestep, as a token, which suits datasets with many correlated channels. DLinear\cite{zeng2023transformers} shows that a simple linear decomposition of trend and seasonal components can match or exceed considerably larger Transformer-based models on several benchmarks. Attention-based multivariate forecasting has also been explored for multi-step prediction; for example, He et al.~\cite{he2023multi} use a multi-attention collaborative network to distinguish the temporal and variable-wise contributions of target and auxiliary series. These architectures, however, operate on structured numerical series and do not explicitly exploit accompanying free-text surveillance information.
	
	\subsection{Multimodal and Text-Augmented forecasting}
	A smaller body of work incorporates auxiliary text into time-series forecasting. In influenza forecasting, Liu et al.~\cite{liu2021forecasting} developed a framework combining influenza observations with Google search queries, using stacked autoencoding for dimensionality reduction and variational mode decomposition before neural forecasting. Their results demonstrate the predictive value of external information sources for influenza surveillance, but the auxiliary information is represented as structured search-query features rather than contextual language representations. TaTS\cite{li2026language} is the main representative of this line: text embeddings are projected into a form compatible with a numerical backbone and combined with it through addition or concatenation, ahead of or within the backbone itself. Approaches of this kind establish a role for text, but they generally do not model an explicit, bidirectional interaction between the two modalities — a text representation may modulate the numerical stream, while the reverse pathway, the numerical trend shaping how the text is read, is typically absent. Cross-modal attention has been effective at modeling this kind of bidirectional interaction in other domains, notably vision-language modeling \cite{lu2019vilbert}, but has seen comparatively little use in time-series forecasting specifically.
	
	\subsection{LLM-based time-series forecasting}
	A separate line of work repurposes pretrained large language models as time-series forecasters. GPT4MTS\cite{jia2024gpt4mts} is the representative baseline here: it reprograms a frozen GPT-2 backbone\cite{radford2019language} for multimodal forecasting by combining patch-based numerical embeddings with a textual prompt. Approaches of this kind benefit from the general-purpose representations learned during large-scale language pretraining, while introducing substantially larger language-model components than purpose-built fusion architectures.
	
	\subsection{ILI and epidemic forecasting}
	Before deep-learning approaches, influenza and epidemic forecasting was commonly formulated through mechanistic transmission models or statistical time-series models. In SIR-family models, the population is represented through compartments such as susceptible, infectious, and recovered individuals, with transmission and recovery parameters governing the movement between compartments; SEIR models additionally introduce an exposed compartment to represent a latent period between infection and infectiousness. These models provide an interpretable description of disease dynamics and can support forecasts of epidemic timing and magnitude, but their performance depends on assumptions about transmission, observation, and population dynamics\cite{shaman2012forecasting}. Statistical approaches such as ARIMA and autoregressive neural-network models instead learn temporal dependence directly from surveillance trajectories and have been used for near-term ILI forecasting\cite{kandula2019near}. 
	
	DSA differs from this body of work along two axes. Relative to unimodal forecasters and traditional disease-dynamics models, it incorporates the contemporaneous textual surveillance signal alongside structured epidemiological measurements. Relative to existing multimodal and LLM-based forecasters, it fuses that information through explicit bidirectional cross-modal attention rather than one-directional conditioning or unconditional concatenation, allowing the numerical and textual streams to shape each other's representation before temporal forecasting.

	\section{Methods}
	
	As discussed in Section 2, existing multimodal approaches to ILI and related time-series forecasting commonly incorporate text by pooling it into a single embedding per time step and concatenating it with a numerical representation, or by using text as a one-way conditioning signal for an otherwise unimodal numerical backbone. These strategies establish a useful role for text but leave the cross-modal interaction itself relatively weakly specified: the textual representation is generally formed independently of the numerical state, while the numerical stream is not explicitly used to determine which parts of the textual signal should matter at a given forecast origin. DSA addresses this gap by keeping the modalities in separate streams and introducing bidirectional Cross-Modal Attention (CMA), so that each stream can condition the representation of the other before temporal aggregation. Section 3.1 formalizes the forecasting task. Section 3.2 then describes DSA's five components in turn — the numerical encoder, the headline encoder, cross-modal attention, the causal temporal model, and the regressor — followed by the training objective.

	\subsection{Problem Formulation}
	At each week $t$, the model observes a numerical feature vector $\mathbf{x}_t \in \mathbb{R}^{14}$ and a paired text snippet $s_t$. Given a lookback window of $T=36$ weeks, $\{(\mathbf{x}_{t-T+1}, s_{t-T+1}), \dots, (\mathbf{x}_t, s_t)\}$, DSA predicts ILI activity for the next $H=12$ weeks, $\mathbf{y} = (y_{t+1}, \dots, y_{t+H}) \in \mathbb{R}^{12}$. This is a direct multi-step forecast: DSA outputs all 12 weeks in a single forward pass rather than predicting one step at a time and feeding it back in. The dataset, splits, and preprocessing used to instantiate $\mathbf{x}_t$ and $s_t$ are described in Section 4.
	
	\subsection{The Dual-Stream Attention Architecture}
	
	\begin{figure*}
		\centering
		\includegraphics[width=\textwidth]{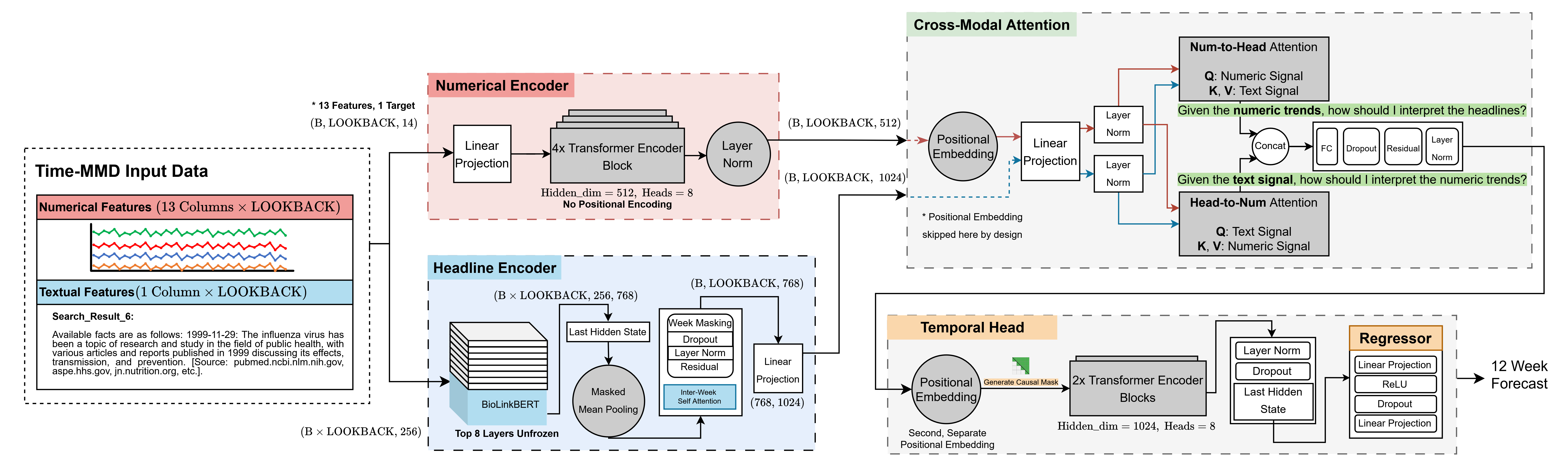}
		\caption{The Dual-Stream Attention (DSA) architecture. The Numerical Encoder and Text Encoder independently represent the 36-week numerical history and its paired weekly text (Sections 3.2.1–3.2.2); Cross-Modal Attention fuses the two streams bidirectionally (Section 3.2.3, Eqs. 4–7); a causal Temporal Model aggregates the fused representation across the lookback window (Section 3.2.4); and a Regressor decodes the pooled representation into the 12-week forecast (Section 3.2.5). Tensor shapes are annotated at each stage.}
		\label{FIG:1}
	\end{figure*}
	
	DSA has five components: a Numerical Encoder and a Headline Encoder, each representing one modality; a CMA module that fuses the two streams; a Causal Temporal Model that aggregates the fused representation across the lookback window; and a Regressor that outputs the forecast. Figure 1 shows the full architecture and the tensor shape at each stage.
	
	\subsubsection{Numerical Encoder Module}
	The numerical stream $\mathbf{X}_{num} \in \mathbb{R}^{T \times 14}$ is projected to the model's hidden size and passed through a 4-layer Transformer encoder\cite{vaswani2017attention}:
	
	\begin{equation}
		\begin{split}
			\mathbf{Z}_{num} ={}& \mathrm{LN}\Bigl(\mathrm{TE}_{\times 4}
			\bigl(\mathbf{X}_{num}W_{in} + \mathbf{b}_{in}\bigr)\Bigr) \in \mathbb{R}^{T \times 512}
		\end{split}
		\tag{1}
	\end{equation}
	
	We do not add positional encoding here. The numerical stream's order is already fixed by the causal mask applied later in the Temporal Model (Section 3.2.4), and adding positional encoding at this stage hurt performance in our experiments (Section 5.4).

	\subsubsection{Headline Encoder Module}
	Each week's text $s_t$ is tokenized (max length $L=256$) and encoded with BioLinkBERT-base\cite{yasunaga2022linkbert}, a language model pretrained on biomedical text. We unfreeze only its last 8 layers during training, which balances adaptation against overfitting at our dataset size. Token representations are combined with masked mean pooling rather than a CLS token, which works better at this scale (Section 5.4):
	
	\begin{equation}
		\mathbf{e}_t = \frac{\sum_{l=1}^{L} m_{t,l}\, \mathbf{h}_{t,l}}{\sum_{l=1}^{L} m_{t,l} + \epsilon}
		\tag{2}
	\end{equation}

	Here $\mathbf{h}_{t,l} \in \mathbb{R}^{768}$ is BioLinkBERT's hidden state for token $l$ of week $t$, and $m_{t,l}\in\{0,1\}$ masks out padding tokens. The resulting weekly embeddings $\mathbf{E}=(\mathbf{e}_1,\dots,\mathbf{e}_T)$ pass through one inter-week self-attention layer, so the text representation at week $t$ can draw on text from other weeks before fusion:
	
	\[
	\mathbf{E}' =
	\mathrm{LN}\Big(
	\mathrm{MHA}(\mathbf{E},\mathbf{E},\mathbf{E})
	+\mathbf{E}
	\Big)
	\tag{3}
	\]
	\[
	\mathrm{MHA}(Q,K,V) =
	\mathrm{softmax}\!\left(
	\frac{QK^\top}{\sqrt{d_k}}
	\right)V
	\]
	
	A dropout layer and a linear projection then map the result into the fusion dimension:
	\[
	\mathbf{Z}_{\mathrm{text}} = \mathrm{Dropout}(\mathbf{E}')\,W_{\mathrm{proj}} \in \mathbb{R}^{T \times 1024}.
	\]
	This text-stream representation, together with the numerical-stream representation $\mathbf{Z}_{num}$ from Eq.~(1), is what Cross-Modal Attention fuses next.

	\subsubsection{Cross-Modal Attention Module}
	
	CMA is DSA's main contribution. Instead of concatenating the two modalities, CMA lets each one condition how the other is read: given the numeric trend, how should the text be interpreted? And given the text, how should the numeric trend be interpreted? Only the text stream gets a learned positional embedding at this stage, matching the choice in Section 3.2.1:
	\[
	\tilde{\mathbf{Z}}_{\mathrm{text}}
	=
	\mathbf{Z}_{\mathrm{text}} + \mathbf{P}_{\mathrm{text}}
	\tag{4}
	\]
	Both streams are projected into a shared dimension and normalized:
	\[
	\hat{\mathbf{Z}}_{\mathrm{num}}
	=
	\mathrm{LN}
	\left(
	\mathbf{Z}_{\mathrm{num}}W_{\mathrm{num}}
	\right)
	\in \mathbb{R}^{T \times 1024}
	\tag{5a}
	\]
	\[
	\hat{\mathbf{Z}}_{\mathrm{text}}
	=
	\mathrm{LN}
	\left(
	\tilde{\mathbf{Z}}_{\mathrm{text}}W_{\mathrm{text}}
	\right)
	\in \mathbb{R}^{T \times 1024}
	\tag{5b}
	\]
	Two attention lanes then run in opposite directions. Their outputs are concatenated, projected back to the fusion dimension, and added to both normalized streams through a residual connection, followed by a final LayerNorm.
	\[
	\mathbf{A}_{n\to t}
	=
	\mathrm{MHA}
	\left(
	Q=\hat{\mathbf{Z}}_{\mathrm{num}},
	K=V=\hat{\mathbf{Z}}_{\mathrm{text}}
	\right)
	\tag{6a}
	\]
	\[
	\mathbf{A}_{t\to n}
	=
	\mathrm{MHA}
	\left(
	Q=\hat{\mathbf{Z}}_{\mathrm{text}},
	K=V=\hat{\mathbf{Z}}_{\mathrm{num}}
	\right)
	\tag{6b}
	\]
	
	\[
	\mathbf{F}
	=
	\mathrm{LN}
	\Big(
	\mathrm{Dropout}
	\big(
	W_{fc}
	[
	\mathbf{A}_{n\to t}
	\,\Vert\,
	\mathbf{A}_{t\to n}
	]
	\big)
	+
	\hat{\mathbf{Z}}_{\mathrm{num}}
	+
	\hat{\mathbf{Z}}_{\mathrm{text}}
	\Big)
	\tag{7}
	\]
	
	We test how faithfully $\mathbf{A}_{n\to t}$ and $\mathbf{A}_{t\to n}$ reflect the model's actual forecasts in Section 5.6, using a perturbation-based masking analysis.
	
	\subsubsection{Temporal Modeling}
	The fused representation receives a second, independent positional embedding, then passes through a 2-layer causal Transformer whose attention is restricted to the current and past positions:
	\[
	\mathbf{F}' = \mathbf{F} + \mathbf{P}_{\mathrm{temporal}},
	\quad
	M_{ij} =
	\begin{cases}
		0, & j \leq i,\\
		-\infty, & j > i
	\end{cases}
	\tag{8}
	\]
	\[
	\mathbf{U}
	=
	\mathrm{LN}
	\Big(
	\mathrm{TE}_{\times 2}
	\big(
	\mathbf{F}',\, \mathrm{mask}=M
	\big)
	\Big)
	\tag{9}
	\]
	Here, $M_{ij}$ is an additive attention mask: a value of $0$ leaves an allowed attention score unchanged, whereas $-\infty$ forces the corresponding softmax attention weight to zero. Thus, $j\leq i$ permits attention to the current and past positions, while $j>i$ blocks access to future positions. The symbol $\mathbf{U}$ denotes the sequence of temporal hidden states produced by the causal Transformer; it is distinct from the scalar $H$, which denotes the forecast horizon.
	
	We take the last position, $\mathbf{u}_{last} = \mathbf{U}_{T}$, as the pooled representation used for forecasting. The causal masking restriction matters: without it, the model can attend to "future" positions within the lookback window during training, which hurts test performance (Section 5.4).
	
	\subsubsection{Regressor and Training Objective}
	
	The regressor is a two-layer multilayer perceptron (MLP) that maps the final temporal representation to the 12-week forecast. DSA is trained end-to-end using mean squared error (MSE) between the predicted and ground-truth ILI values as the training objective:
	\[
	\hat{\mathbf{y}}
	=
	\mathbf{W}_2\,
	\mathrm{Dropout}
	\left(
	\mathrm{ReLU}
	\left(
	\mathbf{W}_1\mathbf{u}_{\mathrm{last}}
	\right)
	\right)
	\in \mathbb{R}^{H},
	\tag{10}
	\]
	where $\mathbf{u}_{\mathrm{last}}$ denotes the final temporal hidden representation, $\mathbf{W}_1$ and $\mathbf{W}_2$ are the learnable MLP weight matrices, and $H$ denotes the forecast horizon, fixed to 12 weeks in this study.
	
	The training objective is defined as
	\[
	\mathcal{L}
	=
	\frac{1}{B}
	\sum_{b=1}^{B}
	\frac{1}{H}
	\sum_{h=1}^{H}
	\left(
	y_{b,h} - \hat{y}_{b,h}
	\right)^2,
	\tag{11}
	\]
	where $B$ is the batch size, $H=12$ is the forecast horizon, $y_{b,h}$ is the ground-truth value for sample $b$ at forecast step $h$, and $\hat{y}_{b,h}$ is the corresponding prediction.
	
	Full training hyperparameters, hardware, and the training-stability investigation that produced the final configuration are reported in Section 4 and Supplementary Appendix A. 
	
	\subsubsection{Algorithmic Summary}
	Algorithm~\ref{alg:dsa} summarizes the full DSA forward pass described in Sections 3.2.1--3.2.5, from the raw numerical history and weekly text to the 12-week forecast, with pointers back to the corresponding equations.
	
	\begin{algorithm}[t]
		\caption{Dual-Stream Attention (DSA) forward pass}
		\label{alg:dsa}
		\begin{algorithmic}[1]
			\Require Numerical history $\mathbf{X}_{num} \in \mathbb{R}^{T\times14}$; weekly text $\{s_1,\dots,s_T\}$ for the same $T=36$ weeks
			\Ensure 12-week forecast $\hat{\mathbf{y}} \in \mathbb{R}^{12}$
			\State $\mathbf{Z}_{num} \gets \mathrm{LN}\big(\mathrm{TE}_{\times 4}(\mathbf{X}_{num}W_{in}+\mathbf{b}_{in})\big)$ \Comment{Numerical Encoder, Eq.~(1)}
			\For{$t = 1$ \textbf{to} $T$}
			\State $\mathbf{h}_{t,1:L} \gets \mathrm{BioLinkBERT}(s_t)$
			\State $\mathbf{e}_t \gets \mathrm{MaskedMeanPool}(\mathbf{h}_{t,1:L})$ \Comment{Eq.~(2)}
			\EndFor
			\State $\mathbf{E} \gets (\mathbf{e}_1,\dots,\mathbf{e}_T)$
			\State $\mathbf{E}' \gets \mathrm{LN}\big(\mathrm{MHA}(\mathbf{E},\mathbf{E},\mathbf{E}) + \mathbf{E}\big)$ \Comment{Inter-week text attention, Eq.~(3)}
			\State $\mathbf{Z}_{text} \gets \mathrm{Dropout}(\mathbf{E}')\,W_{proj}$
			\State $\tilde{\mathbf{Z}}_{text} \gets \mathbf{Z}_{text} + \mathbf{P}_{text}$ \Comment{Eq.~(4)}
			\State $\hat{\mathbf{Z}}_{num} \gets \mathrm{LN}(\mathbf{Z}_{num}W_{num})$; \quad $\hat{\mathbf{Z}}_{text} \gets \mathrm{LN}(\tilde{\mathbf{Z}}_{text}W_{text})$ \Comment{Eq.~(5)}
			\State $\mathbf{A}_{n\to t} \gets \mathrm{MHA}(Q=\hat{\mathbf{Z}}_{num},\,K=V=\hat{\mathbf{Z}}_{text})$ \Comment{Eq.~(6a)}
			\State $\mathbf{A}_{t\to n} \gets \mathrm{MHA}(Q=\hat{\mathbf{Z}}_{text},\,K=V=\hat{\mathbf{Z}}_{num})$ \Comment{Eq.~(6b)}
			\State $\mathbf{F} \gets \mathrm{LN}\Big(\mathrm{Dropout}\big(W_{fc}[\mathbf{A}_{n\to t}\Vert\mathbf{A}_{t\to n}]\big) + \hat{\mathbf{Z}}_{num} + \hat{\mathbf{Z}}_{text}\Big)$ \Comment{CMA fusion, Eq.~(7)}
			\State $\mathbf{F}' \gets \mathbf{F} + \mathbf{P}_{temporal}$
			\State $\mathbf{U} \gets \mathrm{LN}\big(\mathrm{TE}_{\times 2}(\mathbf{F}',\,\mathrm{mask}=M)\big)$ \Comment{Causal Temporal Model, Eqs.~(8)--(9)}
			\State $\mathbf{u}_{last} \gets \mathbf{U}_{T}$
			\State $\hat{\mathbf{y}} \gets W_2\,\mathrm{Dropout}\big(\mathrm{ReLU}(W_1 \mathbf{u}_{last})\big)$ \Comment{Regressor, Eq.~(10)}
			\State \Return $\hat{\mathbf{y}}$
		\end{algorithmic}
	\end{algorithm}
	
	\section{Experiments}
	A fusion architecture is only as convincing as the evaluation behind it. Time-series and multimodal forecasting papers often report a single number from a single train/test split, which makes it hard to tell a real architectural advantage apart from noise in the data or in initialization. We design our experiments to avoid this: multiple random seeds for every core comparison, a broad set of baselines spanning unimodal, naively-fused, purpose-built multimodal, and LLM-based forecasters, and statistical tests that we apply consistently and report them all. This section covers the data (4.1), baselines (4.2), evaluation protocol (4.3), implementation details (4.4), and statistical methodology (4.5) that Section 5's results are built on.
	
	\subsection{Data and preprocessing}
	We evaluate DSA on the Time-MMD Health\_US as the main benchmark\cite{liu2024time} (\texttt{US\_FLURATIO\_Week.csv}), a weekly dataset pairing structured ILI surveillance statistics with real-world text. This dataset is particularly suitable for DSA because the numerical surveillance variables and textual summaries are indexed to the same weekly timeline, providing a natural temporal alignment between the two modalities and avoiding the need for a separate event-matching or cross-modal synchronization procedure. This alignment directly matches DSA's central motivation: to learn how the numerical epidemiological state and the contemporaneous textual context should condition one another at the forecast origin. For the text modality, we use the \texttt{Final\_Search\_6} field, which contains factual, retrieval-based weekly search summaries\cite{liu2024time}. We exclude the alternative \texttt{Final\_Output} field because it contains GPT-3.5-generated predictions, which may introduce information leakage given the model's training-data cutoff\cite{openai2023chatgpt}.

	For the numerical modality, entries marked as \texttt{X} are converted to missing values. Missing age-group values are then reconstructed using the internal consistency of the age categories: when \texttt{AGE 25-64} is available but \texttt{AGE 25-49} and \texttt{AGE 50-64} are missing, the former is partitioned into the two subgroups using a fixed 60/40 split; conversely, when both subgroups are available but their combined category is missing, \texttt{AGE 25-64} is reconstructed as their sum. Numerical features are subsequently scaled using a MinMax scaler fitted exclusively on the training split. Calendar position is encoded using sine and cosine transformations of the ISO week number to provide a continuous seasonal representation across year boundaries. The data are divided chronologically into training, validation, and test sets using an 80/10/10 ratio, ensuring that future test-period information is not used during model development. To assess cross-geographical generalization, DSA is additionally evaluated on the Time-MMD Health\_Africa benchmark in Section~5.5.

	\subsection{Baselines}
	We compare DSA against three families of baselines. Unimodal time-series forecasters — DLinear\cite{zeng2023transformers}, PatchTST\cite{nie2022time}, and iTransformer\cite{liu2024itransformer} — do not use the text modality; they represent a simple linear baseline and two strong recent Transformer-based architectures for long-horizon forecasting. Naive multimodal fusion baselines concatenate a BERT-family text representation with each unimodal backbone's numerical representation before the forecasting head; we use these to test whether multimodal fusion helps at all, independent of how it is done. TaTS \cite{li2026language} is a more purpose-built text-augmented time-series baseline, which we evaluate on each of the three unimodal backbones above. GPT4MTS\cite{jia2024gpt4mts} represents a large-language-model-based forecaster, built on a 6-layer frozen GPT-2 backbone\cite{radford2019language}.
	
	\subsection{Evaluation protocol}
	Evaluating every baseline family across many seeds is expensive, so we use a two-stage protocol. First, we screen all candidate configurations under a single fixed seed to find the strongest representative of each family (full results in Supplementary Appendix B.1). This screening selected four models — DSA, iTransformer, TaTS (iTransformer backbone), and GPT4MTS — for a full evaluation across ten independent seeds (0, 1, 2, 3, 5, 7, 10, 13, 21, 42), which is the basis for every main-text comparison in Section 5.
	
	\begin{figure*}
		\centering
		\includegraphics[width=\textwidth]{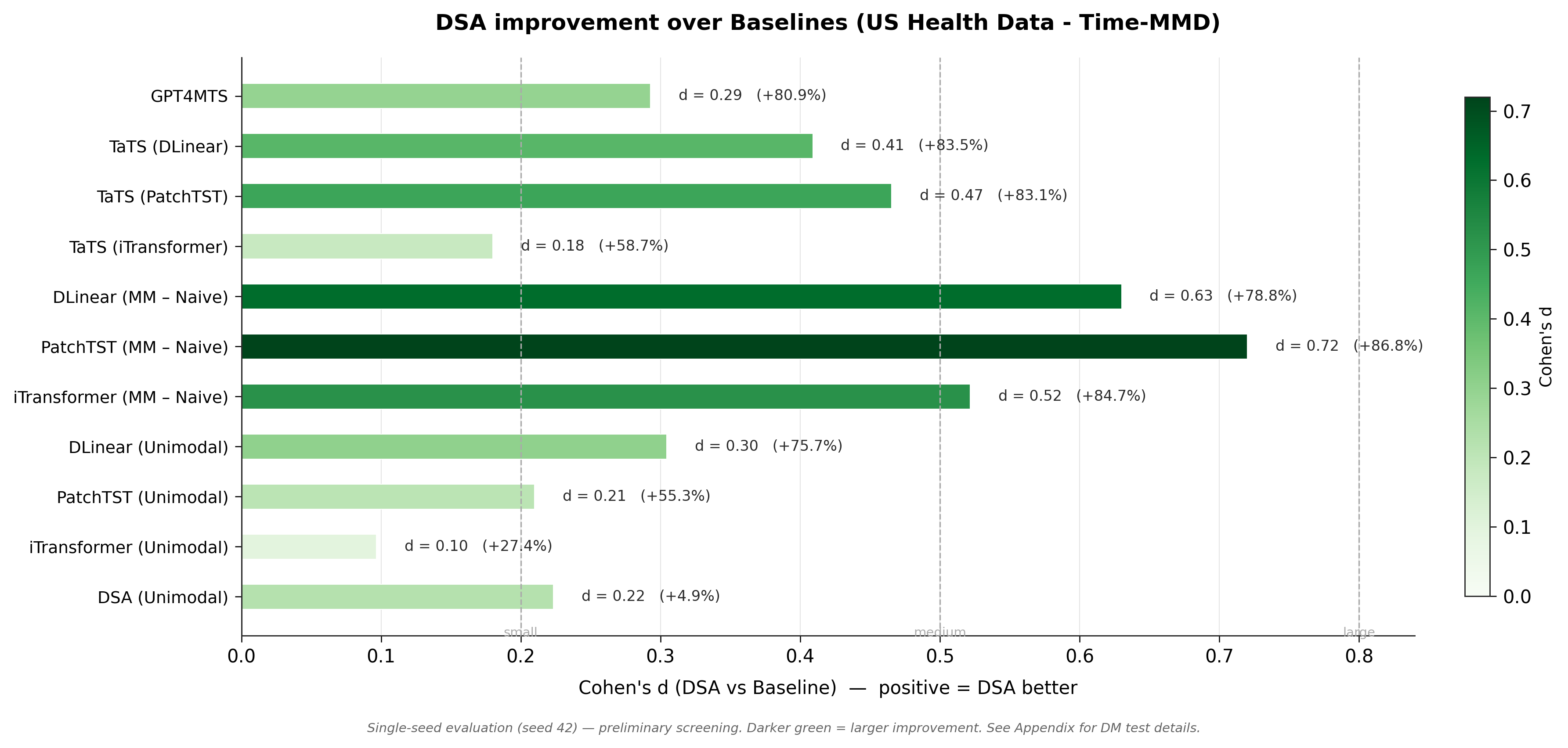}
		\caption{Initial Screening on Time-MMD Health\_US. Effect sizes of DSA over baselines. Positive Cohen’s $d$ values indicate DSA outperforms the respective baseline, with substantial gains over strong competitors like TaTS (iTransformer) ($d=8.42, +98.9\%$). Bar color intensity reflects effect magnitude. (Single-seed evaluation; see Appendix for details.)}
	\end{figure*}
	
	We report median test MSE as the primary metric, along with horizon-averaged MSE, tail-risk (maximum and 90th-percentile per-window MSE), a phase-stratified breakdown by epidemic regime (off-season, rising, peak, declining), and rank-probability / Model Confidence Set analysis (Section 4.5).
	
	\subsection{Implementation details}
	DSA is trained with AdamW\cite{loshchilov2017decoupled} ($\eta=10^{-5}$), dropout $p=0.15$ throughout, and a cosine learning-rate schedule with a 10-epoch warmup excluded from the early-stopping patience counter. We use gradient accumulation over 4 steps (micro-batch size 4, effective batch size 16) under automatic mixed precision. This configuration came out of the reproducibility investigation in Supplementary Appendix A, which traced an early source of seed-to-seed variance to an interaction between the learning-rate scheduler and the accumulation step count. All experiments run on Kaggle's dual-T4 GPU instances. 
	
	\subsection{Statistical methodology}
	Pairwise comparisons use the Diebold–Mariano (DM) test\cite{diebold2002comparing} on the paired loss-differential series $d_\tau = e_{1,\tau}^2 - e_{2,\tau}^2$ between two models' forecast errors on the same test windows:
	\[
	DM = \frac{\bar d}{\sqrt{\widehat{\mathrm{Var}}(\bar d)}}, \qquad \bar d = \frac{1}{N}\sum_{\tau=1}^{N} d_\tau
	\tag{12}
	\]
	which is asymptotically standard normal under the null hypothesis of equal predictive accuracy. $h$-step-ahead forecast errors follow an MA($h$$-$1) process, so the naive lag-0 variance estimator understates the true variance of $\bar d$ at $H=12$. We instead use a Newey–West long-run variance estimator\cite{newey1986simple}:
	\[
	\widehat{\mathrm{Var}}(\bar d)
	=
	\frac{1}{N}
	\left(
	\hat{\gamma}_0
	+
	2\sum_{k=1}^{q}
	\left(
	1-\frac{k}{q+1}
	\right)
	\hat{\gamma}_k
	\right)
	\tag{13}
	\]
	
	\[
	\hat{\gamma}_k
	=
	\frac{1}{N}
	\sum_{\tau=k+1}^{N}
	(d_\tau-\bar d)
	(d_{\tau-k}-\bar d)
	\tag{14}
	\]
	
	with truncation lag $q = H - 1 = 11$.
	
	Effect sizes are reported as paired Cohen's $d$\cite{cohen2013statistical}:
	\[
	d = \frac{\bar d}{s_d}, \qquad s_d = \sqrt{\frac{1}{N-1}\sum_{\tau=1}^{N}(d_\tau - \bar d)^2}
	\tag{15}
	\]
	Percentage improvement is defined as the mean, over the same matched units used for $d_\tau$ in Eq.~(12), of DSA's relative reduction in squared error --- matched per seed for the ten-seed US comparisons, and per test window for the single-run Africa comparison in Section 5.5 --- rather than computed once from the two aggregate MSE values directly:
	\[
	\%\,\mathrm{Improvement} = \frac{1}{N}\sum_{\tau=1}^{N} \frac{e_{baseline,\tau}^2 - e_{DSA,\tau}^2}{e_{baseline,\tau}^2} \times 100\%
	\tag{16}
	\]
	We compute it this way, rather than as $(\mathrm{MSE}_{baseline}-\mathrm{MSE}_{DSA})/\mathrm{MSE}_{baseline}$ applied directly to the two aggregate MSE values, so that it is derived from the same matched pairs as Cohen's $d$ in Eq.~(15) rather than mixing a paired statistic with an unpaired one within the same comparison. Because a mean of per-unit ratios is not generally equal to the ratio of the two means, the two quantities need not agree numerically; Section 5.5 works through a concrete case where they do not, to make this explicit.
	
	We also test directional consistency across the ten seeds with a two-sided exact sign (binomial) test, and corroborate overall ranking with the Model Confidence Set procedure of Hansen, Lunde, and Nason\cite{hansen2011model}, which identifies the subset of models statistically indistinguishable from the best at a given confidence level without relying on any single pairwise test. For the phase-stratified comparisons in Section 5.3, which involve multiple simultaneous tests across regime and baseline pairs, we control the false discovery rate with the Benjamini–Hochberg procedure at $\alpha=0.05$\cite{benjamini1995controlling}.
	
	Per-seed DM testing at this granularity is uncommon in the forecasting literature, and with only ten seeds, individual per-seed tests are comparatively underpowered. We treat the full per-seed workup as a supplementary robustness check (Supplementary Appendix C) rather than the sole basis for our claims. The primary claims in Section 5 rest instead on median MSE, tail-risk, effect size, percentage improvement, directional consistency, and Model Confidence Set inclusion — statistics that do not depend on the choice of variance estimator and are, together, less sensitive to sampling noise at our seed count than any single pairwise test.

	\section{Results \& Discussion}
	
	This section reports the ten-seed comparison motivated in Section 4.3: DSA and its three strongest baselines — iTransformer, TaTS, and GPT4MTS — evaluated jointly on accuracy, tail-risk, and phase-stratified performance rather than on any single metric (Sections 5.1–5.3). We then isolate the source of DSA's advantage through a series of ablations (Section 5.4), test whether it generalizes to a second geography (Section 5.5), and check whether its learned cross-modal attention is functionally meaningful (Section 5.6). Section 5.7 discusses what these results support and where DSA falls short.
	
	\subsection{Main Results}

	According to the results of our experiments (Table~\ref{tab:overall_metrics}), DSA has the best value on all four reported overall metrics: median MSE, horizon-averaged MSE, Max MSE, and P90 MSE. Its horizon-averaged MSE (1.120) is lower than those of TaTS (1.371), GPT4MTS (1.712), and iTransformer (1.753), while its mean maximum-window error (6.404) is also the lowest of the four models. GPT4MTS's horizon-averaged MSE (1.712) sits close to iTransformer's (1.753) — not to TaTS's (1.371), which is the strongest baseline on this metric — while GPT4MTS has the worst median MSE and the highest tail-risk of all four models. This gap between GPT4MTS's average-case and worst-case behavior is examined further below.

	\begin{table*}[width=\textwidth,cols=5,pos=h]
		\caption{Overall error metrics and tail-risk, ten-seed mean $\pm$ s.d. (Max, P90) and median.}
		\label{tab:overall_metrics}
		\small
		\begin{tabular*}{\tblwidth}{@{} L C C C C @{}}
			\toprule
			\textbf{Model} &
			\textbf{Median MSE $\downarrow$} &
			\textbf{Horizon-Avg MSE (mean $\pm$ s.d.)} &
			\textbf{Max MSE (mean $\pm$ s.d.)} &
			\textbf{P90 MSE (mean $\pm$ s.d.)} \\
			\midrule
			DSA (proposed) &
			\textbf{$\mathbf{0.4163}$} &
			\textbf{$\mathbf{1.1196}$ $\pm$ $\mathbf{0.0960}$} &
			$\mathbf{6.4036}$ $\pm$ $\mathbf{0.7903}$ &
			\textbf{$\mathbf{3.1084}$ $\pm$ $\mathbf{0.1981}$} \\
			
			iTransformer &
			$0.6683$ &
			$1.7525$ $\pm$ $0.4874$ &
			$10.7543$ $\pm$ $2.9015$ &
			$4.9619$ $\pm$ $1.6556$ \\
			
			TaTS (iTransformer) &
			$0.6070$ &
			$1.3710$ $\pm$ $0.1771$ &
			$10.1020$ $\pm$ $2.0238$ &
			$3.3136$ $\pm$ $0.7399$ \\
			
			GPT4MTS &
			$0.8519$ &
			$1.7120$ $\pm$ $0.2683$ &
			$14.5867$ $\pm$ $3.3700$ &
			$3.4084$ $\pm$ $0.7180$ \\
			\bottomrule
		\end{tabular*}
	\end{table*}
	
	Furthermore, DSA's advantage is favored by all ten seeds against every baseline, with medium-to-large effect sizes ($d$ = 0.34–0.56) and mean-error improvements of 37–67\% (Table~\ref{tab:effect_size}). Full per-seed Diebold–Mariano statistics are in Supplementary Appendix C.
	
	\begin{table}[width=\linewidth,cols=5,pos=h]
		\caption{Effect size, improvement, and directional consistency of DSA relative to the baseline models.}
		\label{tab:effect_size}
		\small
		\begin{tabular*}{\tblwidth}{@{} L C C C C @{}}
			\toprule
			\textbf{DSA vs} &
			\textbf{\(\Delta \text{MSE}\)} &
			\textbf{$d$} &
			\textbf{\(\text{DSA Imp.} (\%)\)} &
			\textbf{\text{DSA Wins}} \\
			\midrule
			iTransformer &
			$-0.633$ &
			$0.555$ &
			$+54.95\%$ &
			$10/10$ \\
			
			TaTS &
			$-0.251$ &
			$0.337$ &
			$+37.29\%$ &
			$10/10$ \\
			
			GPT4MTS &
			$-0.592$ &
			$0.345$ &
			$+67.23\%$ &
			$10/10$ \\
			\bottomrule
		\end{tabular*}
	\end{table}
	
	\begin{figure*}
		\centering
		\includegraphics[width=\textwidth]{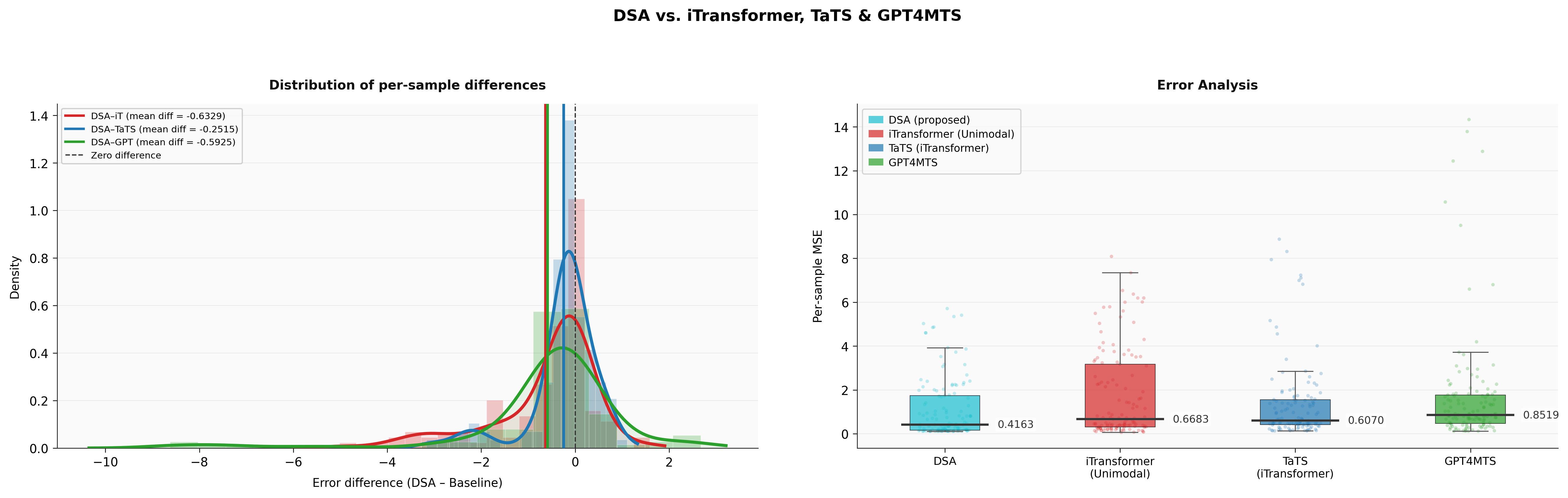}
		\caption{Per-sample error distribution. Left: Density of error differences (DSA -- Baseline); negative values indicate DSA superiority. Right: Per-sample MSE boxplots with annotated means. DSA shows the lowest error and tightest spread, achieving moderate gains over TaTS and substantial improvements over iTransformer and GPT4MTS by avoiding their extreme outliers.}
		\label{FIG:2}
	\end{figure*}

	DSA is ranked first in every bootstrap draw (Table~\ref{tab:rank_probability}). GPT4MTS is ranked last in 74.6\% of draws despite its competitive horizon-averaged error in Table~\ref{tab:overall_metrics} — the clearest illustration of why we do not rely on a single metric. A model with a reasonable average error and the worst tail-risk of the group looks considerably stronger if horizon-averaged MSE is the only number reported.

	\begin{table}[width=\linewidth,cols=5,pos=h]
		\caption{Rank-probability analysis (bootstrap over all four models, ten seeds).}
		\label{tab:rank_probability}
		\small
		\begin{tabular*}{\tblwidth}{@{} L C C C C @{}}
			\toprule
			\textbf{Model} &
			\textbf{Rank 1} &
			\textbf{Rank 2} &
			\textbf{Rank 3} &
			\textbf{Rank 4} \\
			\midrule
			DSA (proposed)  &
			\textbf{$\mathbf{100.0\%}$} &
			$0.0\%$ &
			$0.0\%$ &
			$0.0\%$ \\
			
			TaTS &
			$0.0\%$ &
			$98.7\%$ &
			$1.3\%$ &
			$0.0\%$ \\
			
			iTransformer &
			$0.0\%$ &
			$1.3\%$ &
			$73.2\%$ &
			$25.4\%$ \\
			
			GPT4MTS &
			$0.0\%$ &
			$0.0\%$ &
			$25.4\%$ &
			$74.6\%$ \\
			\bottomrule
		\end{tabular*}
	\end{table}
	
	\begin{figure}
		\centering
		\includegraphics[width=\columnwidth]{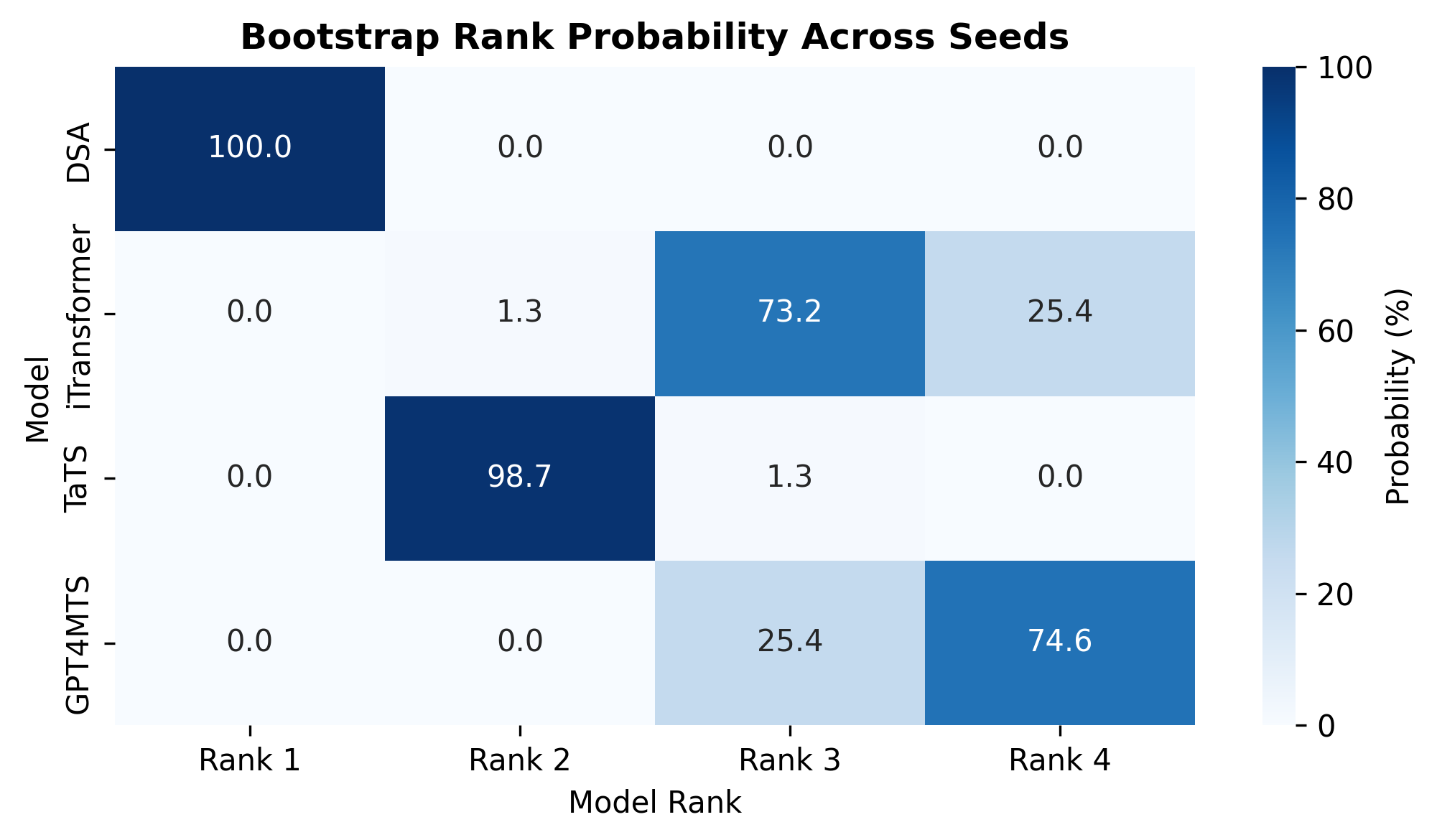}
		\caption{Bootstrap rank probabilities across seeds. DSA ranks first with 100\% probability, followed by TaTS at Rank 2 (98.7\%). iTransformer and GPT4MTS primarily occupy the third and fourth ranks, respectively, corresponding to Table~\ref{tab:rank_probability}.}
		\label{FIG:3}
	\end{figure}

	\subsection{Tail-Risk and Horizon-Wise Robustness}
	
	Tail-risk behavior is relevant for forecasting systems intended to support public health decisions, because inaccurate influenza forecasts can affect downstream decisions such as vaccination strategies, resource allocation, and public communication~\cite{doms2018assessing}. Figure 4 reports the maximum single-window MSE for each of the ten evaluated seeds, providing a direct view of the worst test window encountered by each model under each initialization. Across the ten seeds, the largest observed single-window MSE is 7.703 for DSA, compared with 15.643 for iTransformer, 13.748 for TaTS, and 20.407 for GPT4MTS. Thus, the worst individual seed for DSA remains substantially below the corresponding worst cases of all three baselines.
		
	This per-seed worst-window analysis should be distinguished from the mean maximum-window error reported in Table~\ref{tab:overall_metrics}. Based on the underlying evaluation results, the mean maximum-window errors are 6.4036 for DSA, 10.7543 for iTransformer, 10.1020 for TaTS, and 14.5867 for GPT4MTS. DSA therefore reduces the mean maximum-window error by 4.3507 relative to iTransformer, 3.6984 relative to TaTS, and 8.1831 relative to GPT4MTS. Full per-seed tail-risk results are provided in Supplementary Appendix~\ref{app:tail_horizon}. Together, these results indicate that the advantage of DSA is not limited to its average test performance but also extends to the most difficult forecasting windows across random initializations.
	
	\begin{figure}
		\centering
		\includegraphics[width=\columnwidth]{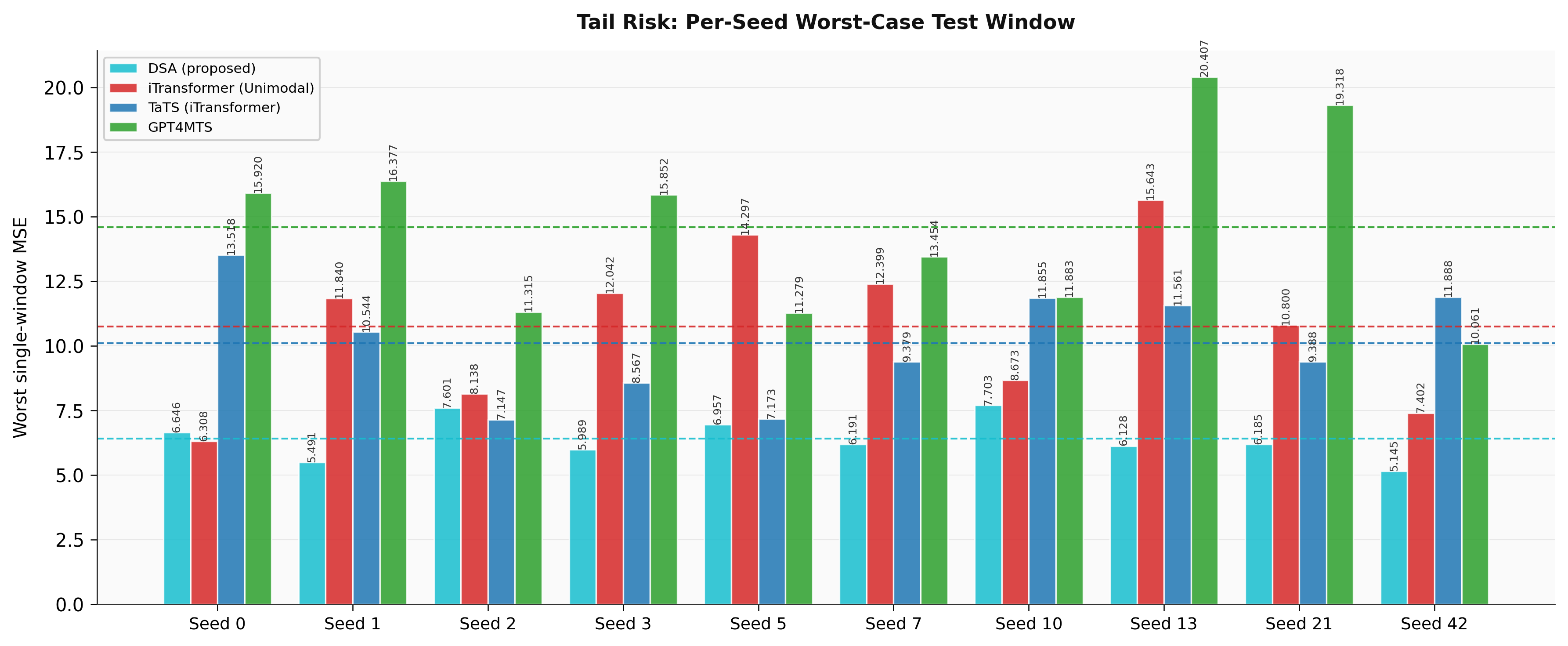}
		\caption{Tail-risk comparison (worst-window and P90 MSE, ten-seed distribution) for DSA against baselines, corresponding to the Max and P90 columns of Table~\ref{tab:overall_metrics}. Notably, DSA achieves the lowest average worst-case error compared with baselines, whereas GPT4MTS shows the highest tail risk.}
		\label{FIG:4}
	\end{figure}
	
	Horizon-wise (Figure 6; per-horizon detail in Figure 7), DSA's advantage over iTransformer grows with the forecast horizon, from near parity at horizon 1 to its largest gap at horizon 10. Its advantage over TaTS peaks mid-horizon, around week 7, and its advantage over GPT4MTS peaks later, around week 11. We do not observe a horizon at which any baseline outperforms DSA on average, though the margin is narrow near horizon 1 in all three comparisons.
	
	\begin{figure*}
		\centering
		\includegraphics[width=\textwidth]{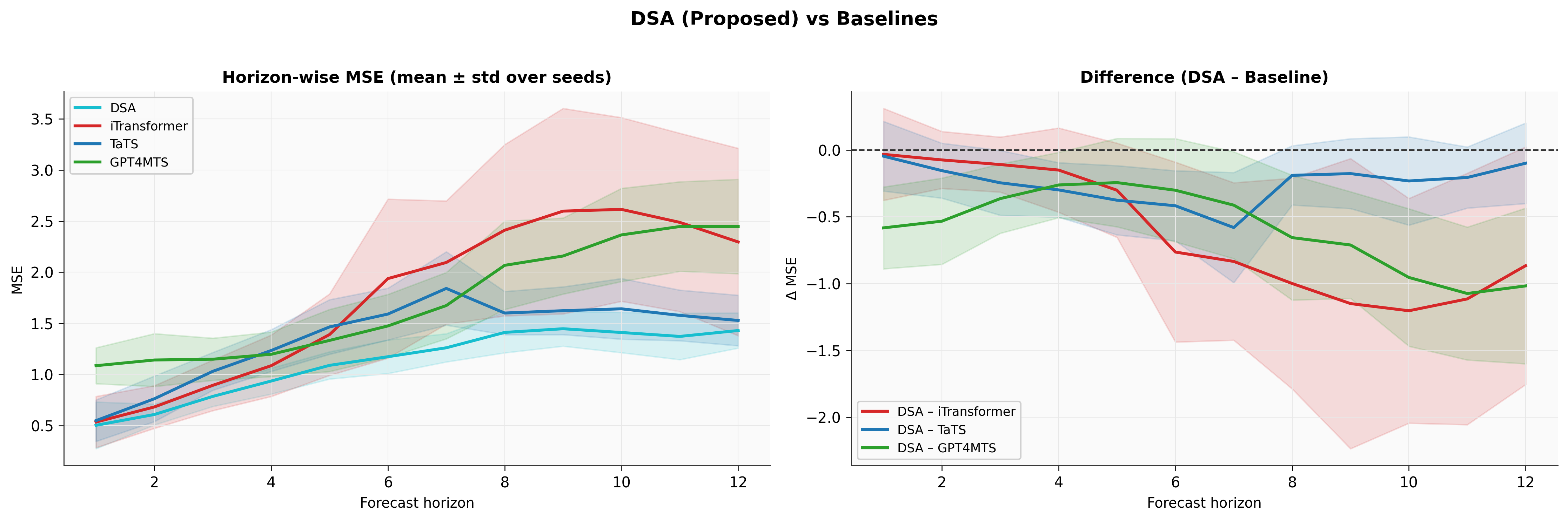}
		\caption{Forecasting performance of DSA versus baselines. (Left) Horizon-wise MSE (mean \(\pm\) std over seeds), illustrating DSA's lower absolute error and tighter variance across the forecast horizon. (Right) MSE difference (\(\Delta\) MSE = DSA \(-\) Baseline); the dashed zero-line acts as a threshold where negative values indicate DSA outperforms the respective model by that shaded margin. DSA consistently maintains a negative \(\Delta\) MSE against all baselines nearly over the entire horizon.}
		\label{FIG:5}
	\end{figure*}
	
	\begin{figure*}
		\centering
		\includegraphics[width=\textwidth]{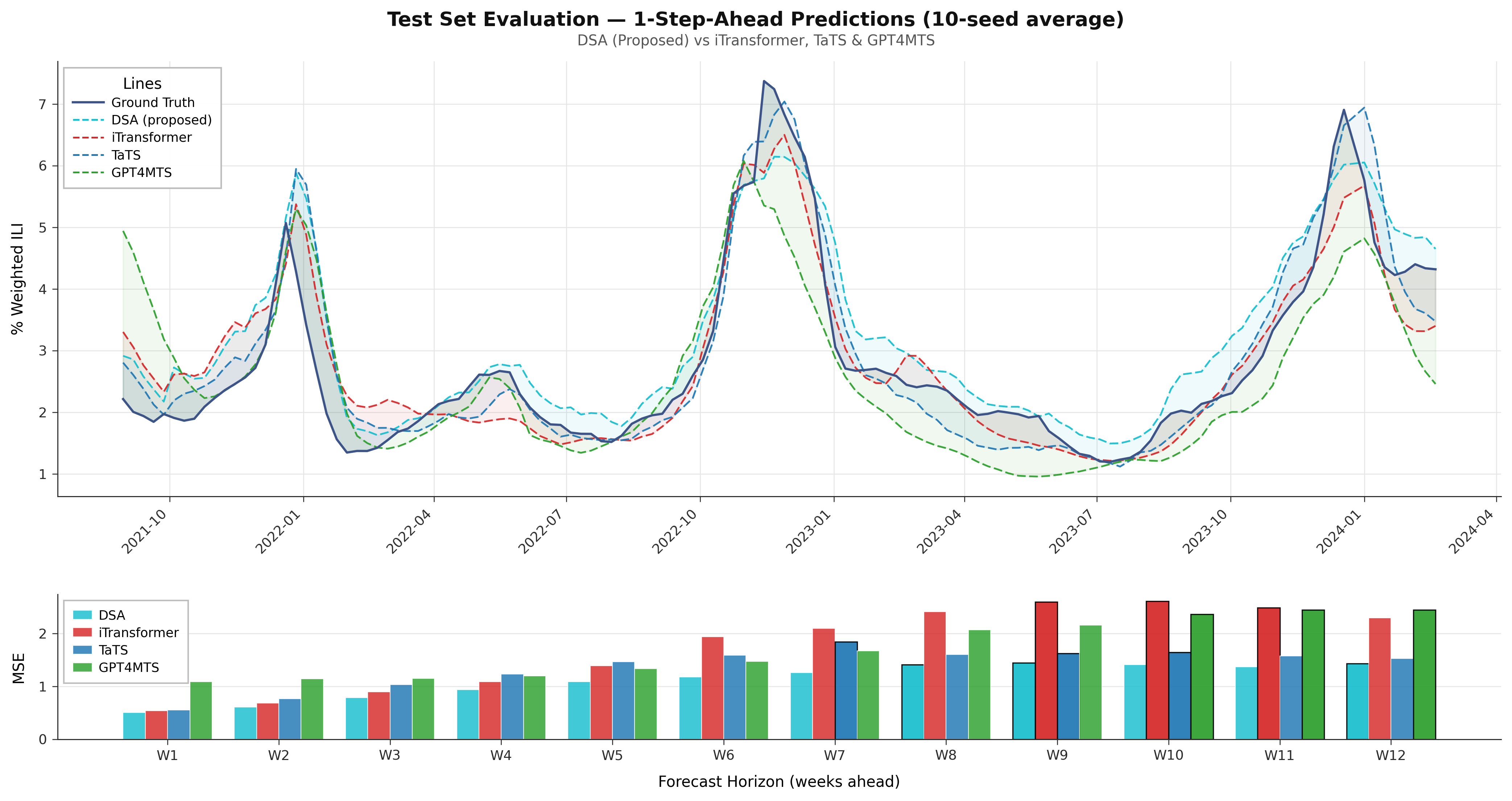}
		\caption{Per-horizon test MSE for DSA, iTransformer, TaTS, and GPT4MTS across forecast weeks 1--12, showing the near-parity at horizon 1 and the growing advantage at later horizons described in Section 5.2.}
		\label{FIG:6}
	\end{figure*}
	
	\subsection{Phase-Stratified Robustness}
	 Aggregate metrics can hide regime-specific weaknesses, which matters for a seasonal disease like influenza\cite{kim2024predicting}. We split test windows into four epidemic phases — off-season, rising, peak, and declining — and repeat the comparison within each; full numeric values for this section are in Supplementary Table~\ref{tab:supp_phase}.
	
	\begin{figure}
		\centering
		\includegraphics[width=\columnwidth]{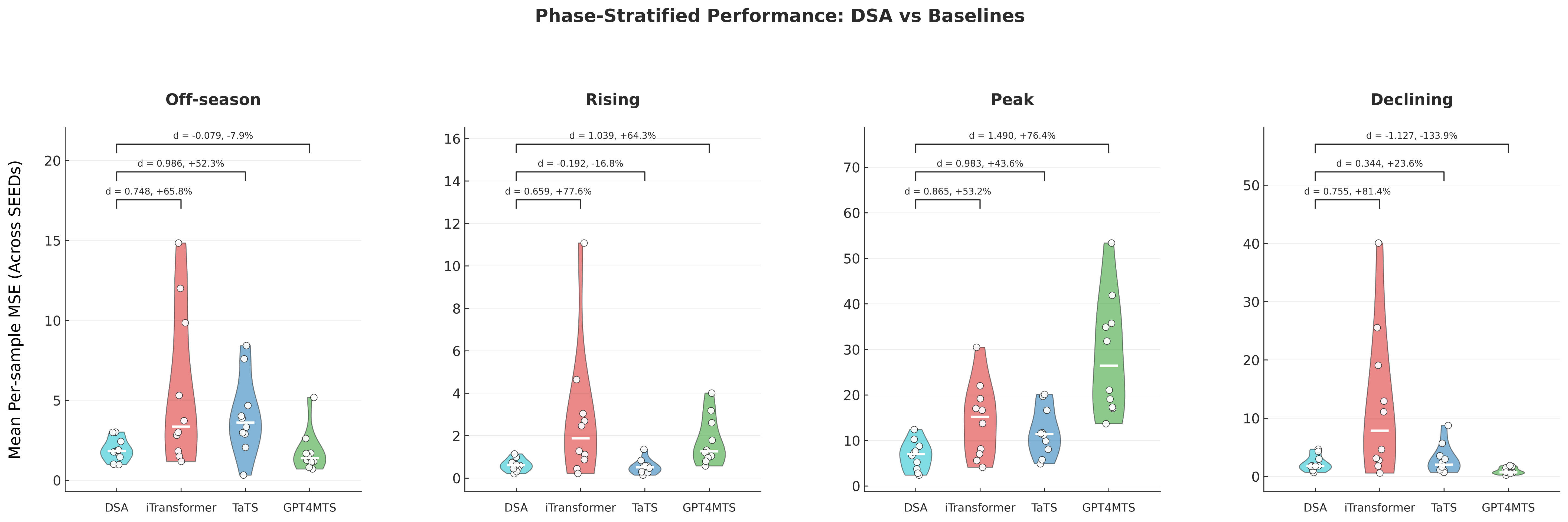}
		\caption{Phase-stratified mean per-sample MSE. Notably, while DSA is the top-performing model during the critical Peak phase and remains highly competitive in Off-season and Rising phases, it is outperformed by GPT4MTS during the Declining phase.}
		\label{FIG:7}
	\end{figure}
	
	DSA has the lowest mean MSE in three of the four phases (off-season, rising, peak; Supplementary Table~\ref{tab:supp_phase}). Two exceptions are worth stating plainly rather than folding into the aggregate result. First, DSA underperforms TaTS in the rising phase (Cohen's $d = -0.19$; TaTS's mean MSE is 16.8\% lower). Second, DSA underperforms GPT4MTS in two phases: marginally in the off-season ($d = -0.08$) and substantially in the declining phase ($d = -1.13$), where GPT4MTS's mean MSE (0.97) is less than half of DSA's (2.26). Model Confidence Set inclusion, corrected within each phase, is consistent with this picture: DSA remains in the confidence set at all three thresholds we test (75\%/90\%/95\%) in the off-season, rising, and peak phases, but only at the two more permissive thresholds in the declining phase.
	
	These exceptions do not, on their own, undercut the case for DSA. Averaged across all four phases, DSA still has the lowest mean MSE of the four models, and it does so while also being the clear best model in the peak phase (Figure~\ref{FIG:7}; Supplementary Table~\ref{tab:supp_phase}) — the regime with the largest errors for every model and the one operationally most consequential, since peak-season forecasts are what public-health resource allocation depends on most. In the two phases where it is not the single best model, the gap is comparatively small: DSA trails TaTS in the rising phase by $d=-0.19$ and GPT4MTS in the off-season by $d=-0.08$, and it remains inside the Model Confidence Set at every threshold tested in both. Only in the declining phase, where GPT4MTS pulls decisively ahead ($d=-1.13$), does DSA's disadvantage become substantial. We read this pattern as DSA being consistently competitive rather than uniformly dominant — a property that matters in deployment, where a forecasting system must produce an estimate every week without first knowing which phase that week belongs to.
	
	\begin{figure}
		\centering
		\includegraphics[width=\columnwidth]{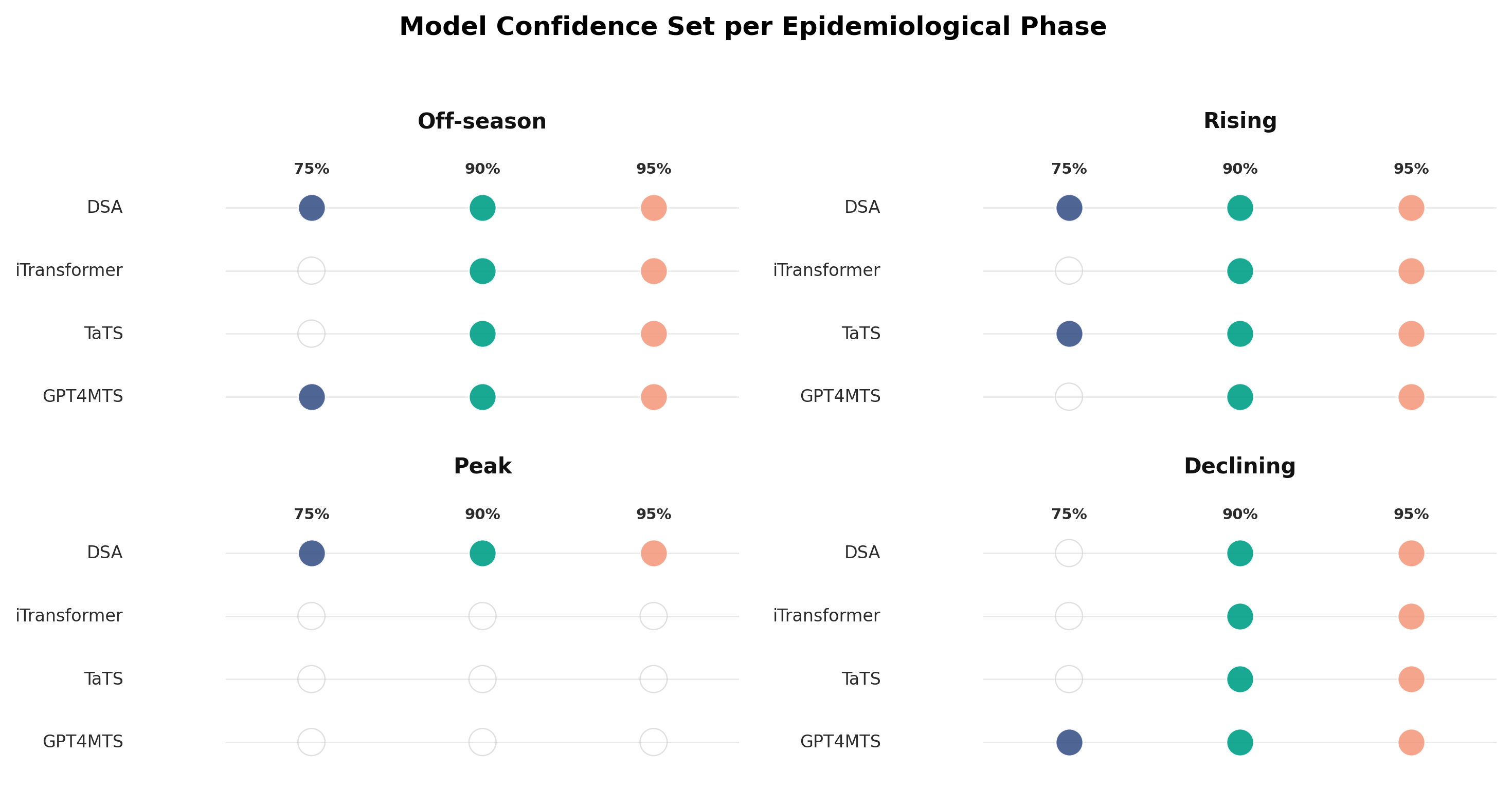}
		\caption{Model Confidence Set per Epidemiological Phase. Filled circles denote models included in the statistical confidence set at 75\%, 90\%, and 95\% levels. While DSA is the sole model retained throughout the critical Peak phase, it remains on par with competing baselines across the Off-season, Rising, and Declining phases, corresponding to the phase-level thresholds discussed in Section 5.3.}
		\label{FIG:8}
	\end{figure}
	
	\subsection{Ablation Studies}
	We isolate the contribution of each design choice through four sets of ablations:
	an architecture search over individual components, a test of whether bidirectional
	attention itself is necessary, a test of whether fusion or fine-tuning drives DSA's
	advantage, and a test of sensitivity to the text encoder. The architecture-search
	results are summarized in Table~\ref{tab:supp_architecture} in Supplementary Appendix B.2.

	The causal mask (Section 3.2.4) is the single most load-bearing component we tested: removing it raises test MSE from 1.09 to 1.71. Reversible instance normalization (RevIN)\cite{kim2021reversible}, a common preprocessing step for non-stationary series, degrades performance here (1.80); we attribute this to ILI curves being non-stationary in a way that makes lookback-window statistics a poor guide to the test period. Mean pooling outperforms a CLS-token representation at this data scale (1.09 vs. 1.28), and replacing the numerical encoder with an iTransformer-style design is worse across the board (1.83).
	
	To test whether DSA's advantage comes from fine-tuning BioLinkBERT or from the fusion mechanism itself, we freeze the text encoder entirely and retrain the rest of the model. The frozen-encoder variant still achieves a lower ten-seed mean MSE (1.217 ± 0.194) than the strongest fusion baseline, TaTS (1.371 ± 0.177), even though it cannot adapt its language representations to the task (full detail in Supplementary Appendix~\ref{app:frozen_encoder}). This indicates the fusion mechanism, not fine-tuning, accounts for most of DSA's advantage over TaTS; fine-tuning provides a further improvement (full DSA: 1.120 ± 0.096) but is not the primary source of the gain.
	
	DSA's central design choice is bidirectional CMA rather than a single attention direction. To test whether this earns its added complexity, we compare full DSA against two ablated variants that each retain only one attention branch, with a properly sized fusion projection rather than bidirectional's parameter budget with one branch dropped: Text to Numerical only (numerical queries text) and Numerical to Text only (text queries numerical). At a single seed, matching the fixed-seed architecture-search protocol of Section 4.3, bidirectional DSA achieves the lowest test MSE (1.0367, MAE 0.7131, RMSE 1.0182), ahead of the Text$\rightarrow$Numerical-only variant (MSE 1.0971, MAE 0.7534, RMSE 1.0474) and the Numerical$\rightarrow$Text-only variant (MSE 1.1508, MAE 0.7821, RMSE 1.0728; full detail in Supplementary Appendix~\ref{app:directional_cma}). This ordering is consistent with the faithfulness asymmetry reported in Section 5.6: the direction found to be more informative under perturbation also degrades DSA the least when the other direction is removed, while dropping the more informative direction costs more. At this single seed, neither direction is redundant; we have not repeated this ablation across multiple seeds, so we report it as a single-run directional result rather than a statistically validated one.
	
	A separate concern is whether DSA depends on an expensive, domain-adapted language model, and if so, whether that dependence comes from BioLinkBERT's biomedical pretraining or simply from its parameter count. We repeat the pipeline with two general-domain alternatives that isolate these two factors: BERT-base-uncased, which matches BioLinkBERT's parameter count but lacks biomedical pretraining, tests how much the domain adaptation itself contributes; MiniLM-L6-v2, at roughly one-fifth the parameters, tests how much encoder capacity matters. Both are evaluated across five seeds.
	
	BioLinkBERT gives the best median MSE of the three, and the ordering is consistent on tail-risk (Max and P90 MSE; Figure~\ref{FIG:9}; full numeric values in Supplementary Table~\ref{tab:supp_text_encoders}). The gap, however, is not large enough to suggest DSA collapses without BioLinkBERT specifically: both general-domain substitutes cost some accuracy but leave the qualitative picture unchanged, indicating that DSA's advantage depends on the fusion mechanism more than on either the biomedical pretraining or the raw capacity of any single text encoder.
	
	\begin{figure*}
		\centering
		\includegraphics[width=\textwidth]{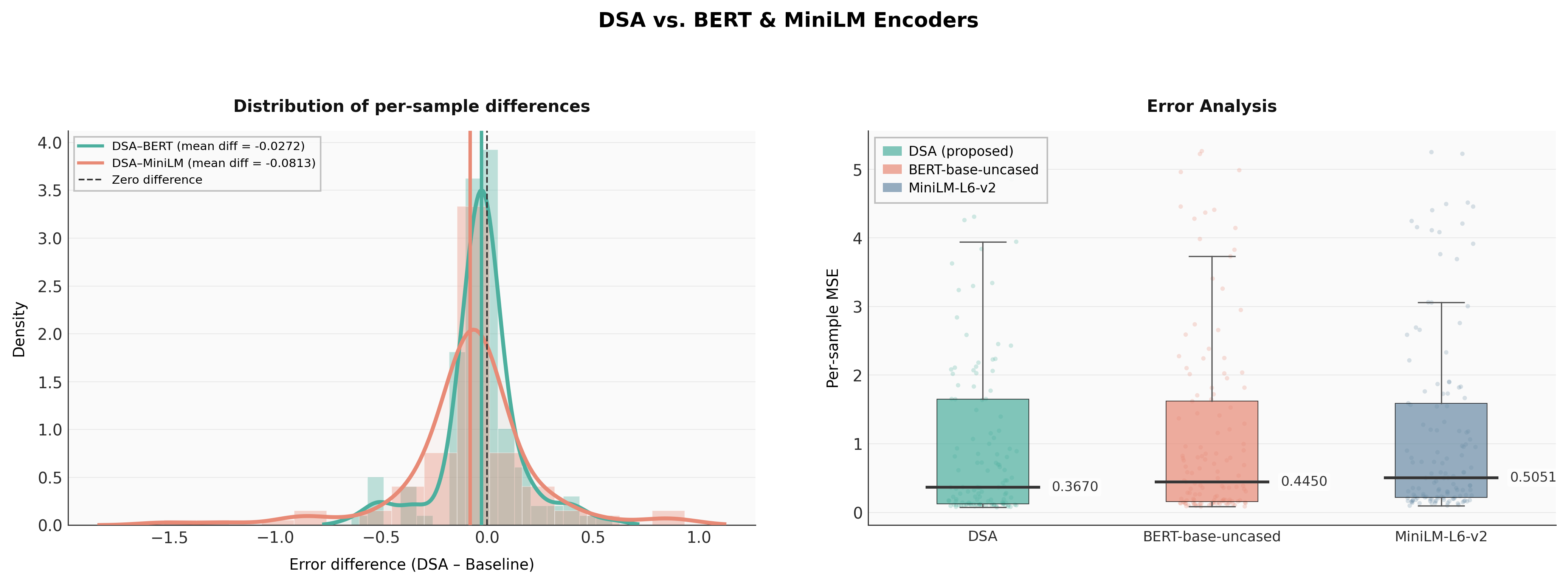}
		\caption{Encoder ablation. (Left) Negligible error differences between BioLinkBERT and other encoders (-0.027, -0.081). (Right) Closely matched MSE distributions confirm CMA architecture (not encoder richness) drives performance. See Supplementary Appendix~\ref{app:text_encoder} for tail-risk analysis.}
		\label{FIG:9}
	\end{figure*}
	
	\subsection{Generalization to a Second Geography}
	If DSA's advantage were specific to this dataset's particular text–numerical relationship, we would not expect it to transfer to a different geography.
	
	\begin{table*}[t]
		\caption{External validation on Health\_Africa (single run per model).}
		\label{tab:health_africa}
		\small
		\begin{tabular*}{\tblwidth}{@{} L C C C C C @{}}
			\toprule
			\textbf{Model} &
			\textbf{Test MSE} &
			\textbf{MAE} &
			\textbf{RMSE} &
			\textbf{$d$} &
			\textbf{\% DSA Imp.(\%)} \\
			\midrule
			DSA (proposed) &
			\textbf{$\mathbf{0.0040}$} &
			\textbf{$\mathbf{0.0543}$} &
			\textbf{$\mathbf{0.0632}$} &
			--- &
			--- \\
			
			GPT4MTS &
			$0.0050$ &
			$0.0589$ &
			$0.0704$ &
			$0.68$ &
			$34.75\%$ \\
			
			DSA (Unimodal) &
			$0.0053$ &
			$0.0556$ &
			$0.0725$ &
			$0.57$ &
			$43.24\%$ \\
			
			DLinear (Unimodal) &
			$0.0086$ &
			$0.0817$ &
			$0.0930$ &
			$1.96$ &
			$78.44\%$ \\
			
			TaTS + DLinear &
			$0.0140$ &
			$0.1116$ &
			$0.1183$ &
			$3.11$ &
			$91.55\%$ \\
			
			PatchTST (Unimodal) &
			$0.0175$ &
			$0.1069$ &
			$0.1321$ &
			$4.69$ &
			$94.23\%$ \\
			
			iTransformer &
			$0.0254$ &
			$0.1375$ &
			$0.1592$ &
			$3.02$ &
			$97.40\%$ \\
			
			TaTS + PatchTST &
			$0.0279$ &
			$0.1329$ &
			$0.1670$ &
			$6.43$ &
			$97.75\%$ \\
			
			TaTS + iTransformer &
			$0.0407$ &
			$0.1709$ &
			$0.2018$ &
			$8.42$ &
			$98.94\%$ \\
			\bottomrule
		\end{tabular*}
		\vspace{2pt}
		{\footnotesize $^{\dagger}$Mean of the per-window \% reductions in squared error (Eq.~16), computed from the same paired per-window differentials as Cohen's $d$ --- not the \% change between the two aggregate MSE values shown in this table. See the paragraph below and Supplementary Appendix~\ref{app:africa_error} for the reference calculation worked out in full.}
	\end{table*}
	
	The \% Improvement and Cohen's $d$ columns above are paired statistics computed per test window (Eqs.~15--16) and then averaged; they are not derived from the two aggregate MSE values shown in the first column pair. 
	We report the paired version so that \% Improvement and Cohen's $d$ in the same row come from the same underlying per-window differentials --- consistent with the pairing used throughout Section 4.5 --- rather than mixing a paired effect-size statistic with an unpaired aggregate ratio in the same table. The direct, unpaired reduction computed from the two aggregate MSE values for every model in this table --- the quantity a reader would otherwise compute by hand from the columns shown here --- is worked out explicitly in Supplementary Appendix~\ref{app:africa_error}.
	
	DSA (multimodal) remains best on all three error metrics (Table~\ref{tab:health_africa}). GPT4MTS is the strongest baseline here, consistent with its improved configuration on Health\_US. TaTS underperforms its own unimodal backbones on this dataset — a pattern absent on Health\_US — even after a supplementary tuning investigation (Supplementary Appendix~\ref{app:africa_tuning}) that increased the text-to-numerical dimension mapping from 12 to 32 and left results essentially unchanged or worse across all three backbones. This suggests weaker text-fusion strategies can fail to generalize across geographies in a way DSA's cross-modal attention does not, though this is a single external dataset, and a broader multi-region evaluation would be needed to treat this as a general property of the architecture.
	
	\begin{figure*}
		\centering
		\includegraphics[width=\textwidth]{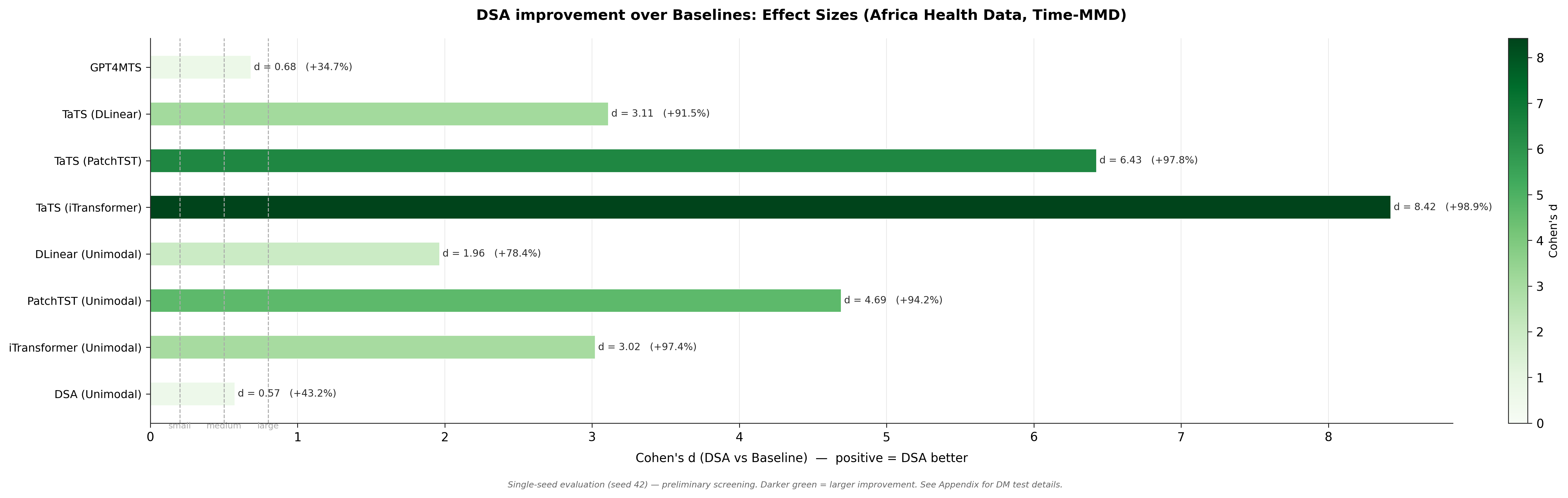}
		\caption{External validation on Time-MMD Health\_Africa. Effect sizes of DSA over baselines. Positive Cohen’s $d$ values indicate DSA outperforms the respective baseline, with substantial gains over strong competitors like TaTS (iTransformer) ($d=8.42, +98.9\%$). Bar color intensity reflects effect magnitude. (Single-seed evaluation; see Appendix for details.)}
		\label{FIG:10}
	\end{figure*}
	
	\subsection{Interpretability: CMA Faithfulness}
	CMA's attention weights (Eq. 6) are useful as an explanation only if they reflect what the model actually relies on. We test this with a perturbation analysis: for each direction (Numerical→Text and Text→Numerical), we mask the top-$k$, a random-$k$, or the bottom-$k$ attended weeks of the lookback window and measure the resulting change in test MSE. If attention is functionally informative, masking the top-attended weeks should hurt the forecast more than masking random or weakly-attended ones.
	
	\begin{table}[width=\linewidth,cols=6,pos=h]
		\caption{Perturbation effect at $k=3$ measured by $\Delta$MSE.}
		\label{tab:perturbation_effect}
		\small
		\begin{tabular*}{\tblwidth}{@{} L C C C C C @{}}
			\toprule
			\textbf{Direction} &
			\textbf{Top-3} &
			\textbf{Rand-3} &
			\textbf{Bottom-3} &
			\textbf{$p$-value} &
			\textbf{$d$} \\
			\midrule
			Num $\rightarrow$ Text &
			$0.0352$ &
			$0.0199$ &
			$0.0171$ &
			$0.0126$ &
			$0.243$ \\
			
			Text $\rightarrow$ Num &
			$1.4436$ &
			$0.1665$ &
			$0.0025$ &
			$2.06\times10^{-7}$ &
			$0.411$ \\
			\bottomrule
		\end{tabular*}
	\end{table}
	
	Both directions show the expected ordering — Top-$k$ $>$ Random-$k$ $>$ Bottom-$k$ (Table~\ref{tab:perturbation_effect}) — and the effect holds across masking budgets from $k$=1 to $k$=12 (2.8\%–33.3\% of the lookback window), widening as more weeks are masked (Supplementary Appendix~\ref{app:faithfulness}). The effect is substantially stronger and more consistent in the Text→Numerical direction: masking the top-3 attended weeks raises MSE by 1.44 on average, roughly 40 times the corresponding effect in the Numerical→Text direction (0.035).
	
	We read these results as evidence that DSA's cross-modal attention is functionally informative under targeted perturbation, rather than as evidence that attention is causal or that every highly-attended week is individually necessary to the forecast; establishing a genuinely causal claim would require a richer, more temporally diverse dataset and a broader set of controlled interventions than we run here. We return to this distinction, and to the recency confound that further limits a causal reading of the Text$\rightarrow$Numerical result, in Section 5.7.
	
	As a complementary check, we also compute global feature-importance scores with Integrated Gradients\cite{sundararajan2017axiomatic}, attributing DSA's output to individual input features across both streams rather than to whole lookback weeks. 
	
	\begin{figure}
		\centering
		\includegraphics[width=\columnwidth]{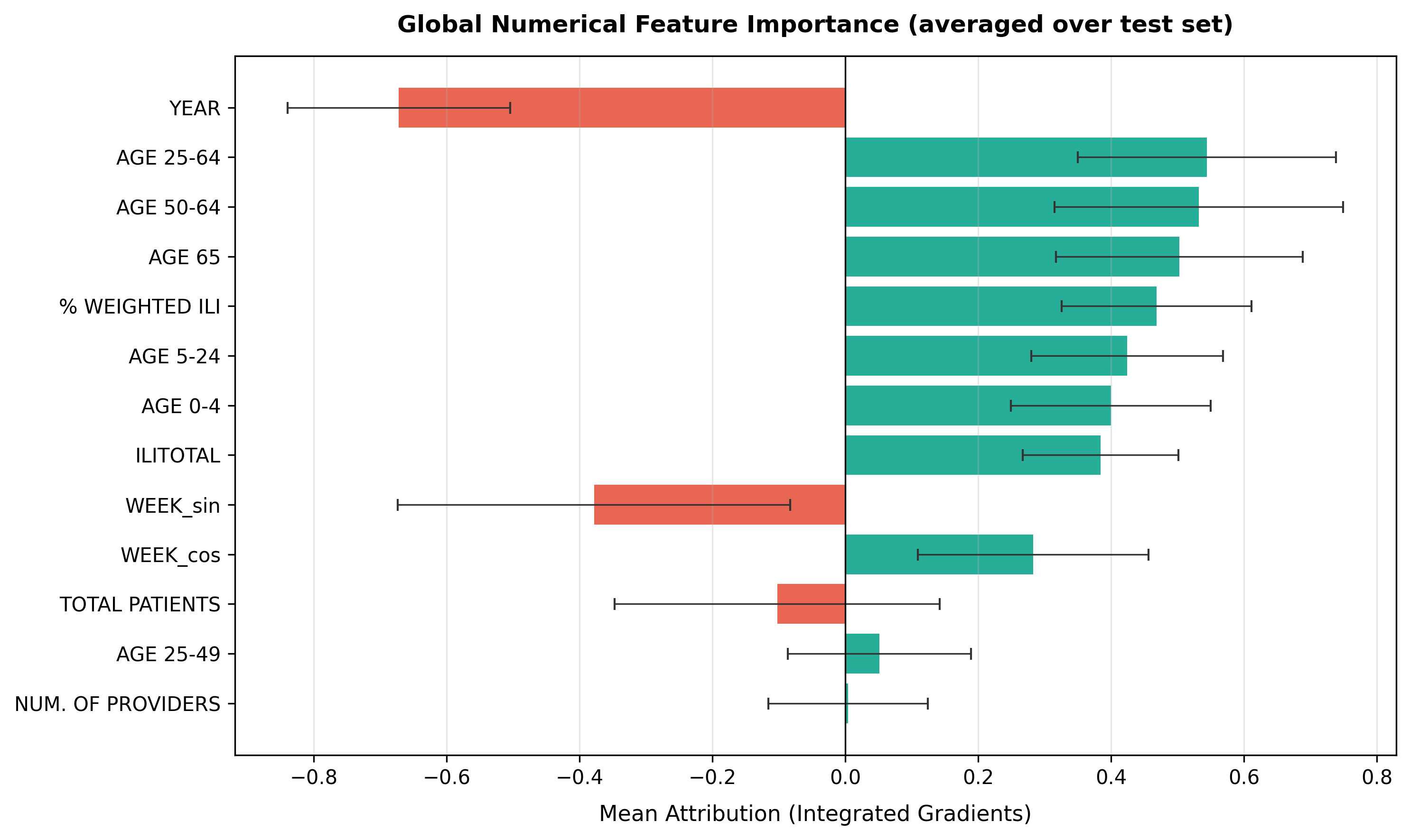}
		\caption{Global feature-importance scores from Integrated Gradients, attributing DSA's forecasts to individual numerical and text-derived input features.}
		\label{FIG:11}
	\end{figure}
	
	\subsection{Discussion and Limitations}
	Taken together, these results support three claims, each grounded in a different piece of evidence rather than in the aggregate accuracy numbers alone. \textbf{First}, cross-modal fusion — modeling how each modality should condition the interpretation of the other — is what makes text useful here, not merely the presence of a second modality. Two results support this specifically, from opposite directions. Naive fusion baselines, which concatenate a pooled text embedding with a numerical backbone's representation without any conditioning mechanism, were severely degraded relative to their corresponding unimodal backbones in our screening pass (Supplementary Appendix B.1) — text hurt more than it helped once it was added without a fusion mechanism. Conversely, the frozen-encoder ablation (Section 5.4) shows that even with BioLinkBERT's parameters entirely frozen, DSA's bidirectional CMA still outperforms the strongest fusion baseline, TaTS (1.217 $\pm$ 0.194 vs.\ 1.371 $\pm$ 0.177 ten-seed mean MSE); full fine-tuning narrows this further (1.120 $\pm$ 0.096) but is not what separates DSA from TaTS. Together, these two results triangulate the same conclusion from opposite ends: neither adding text without conditioning, nor conditioning without fine-tuning, is sufficient on its own — it is the fusion mechanism specifically that drives the advantage. A single-seed ablation of the attention mechanism itself points in the same direction: retaining only one CMA direction, Text$\rightarrow$Numerical or Numerical$\rightarrow$Text, degrades accuracy relative to full bidirectional DSA in both cases (Section 5.4), consistent with bidirectionality contributing rather than being redundant, though we have not validated this specific comparison across multiple seeds as we did for the other core results. \textbf{Second}, DSA's advantage is not confined to average-case accuracy: it holds under tail-risk, phase-stratified, and cross-geography evaluation, which matters more for a deployed forecasting system than any single aggregate metric would suggest. \textbf{Third}, the mechanism responsible for this advantage is not just a design choice that happens to improve accuracy — its learned attention weights are functionally informative under direct perturbation testing, which is stronger evidence than accuracy alone.
	
	These results also come with real qualifications, which we state directly rather than fold into the aggregate narrative. While DSA outperforms each rival baseline on average, it does not dominate every baseline in every regime once the comparison is stratified by phase: it underperforms TaTS in the rising phase and GPT4MTS in the off-season and, substantially, in the declining phase (Section 5.3). Statistical significance at the per-seed level is also more modest than the aggregate picture alone might suggest: the Newey--West correction for serial dependence in the forecast-error differential is, by design, more conservative than an uncorrected variance estimate, and under it only the DSA-versus-iTransformer comparison reaches conventional significance at individual seeds ($p=0.025$), while DSA-versus-TaTS and DSA-versus-GPT4MTS do not ($p=0.106$ and $p=0.142$; full results in Supplementary Appendix~\ref{app:statistics}). We read this as an expected consequence of a conservative correction rather than as evidence against the effect: the direction of the comparison, the effect sizes, the bootstrap rank probabilities, and the Model Confidence Set results in Tables~\ref{tab:effect_size} and~\ref{tab:rank_probability} are all mutually consistent, and it is this convergence across independent statistical lenses, rather than per-seed significance in isolation, that we treat as our primary evidence for DSA's advantage. The Text$\rightarrow$Numerical perturbation result (Section 5.6) also correlates with recency — the most heavily attended weeks in that direction tend to be the most recent ones in the lookback window. This is not entirely unexpected for weekly surveillance narrative, which by nature describes the current epidemiological state rather than a retrospective account, so some of that attention weight plausibly reflects genuine informativeness rather than a pure artifact; even so, the present dataset does not let us fully separate the two, so we flag it as a recency confound that tempers a purely causal reading of this result (full discussion in Supplementary Appendix~\ref{app:faithfulness}). Finally, DSA pools the lookback window into a single fixed-size representation before decoding the forecast (Section 3.2.4); this becomes a bottleneck at longer horizons, and we did not find a configuration that extends cleanly beyond $H=12$ without a richer decoding mechanism than the one used here.

	\section{Conclusion}

	This study introduced Dual-Stream Attention (DSA), a multimodal framework for 12-week ILI forecasting that explicitly models interactions between a 36-week numerical epidemiological history and concurrent weekly text. Its central design choice is bidirectional Cross-Modal Attention: rather than appending text to a numerical forecast representation, DSA allows the numerical and textual streams to condition one another before causal temporal aggregation and direct multi-step decoding. This yields a simple, communicable inductive bias: the numerical history should inform how the text is read, and the text should inform how the numerical history is interpreted.
	
	Across ten independent seeds on Time-MMD Health\_US, DSA provides the strongest overall error profile among the top four models, with the lowest median MSE and markedly lower tail-risk; it also attains the highest bootstrap rank-1 probability among these models. Its advantage is reflected in consistent direction across all ten seeds and favorable effect sizes against iTransformer, TaTS, and GPT4MTS.
	
	The phase-stratified results deserve particular attention. DSA is not always the single best model in every epidemiological phase, but it is consistently among the top two in all four phases. This is the more relevant property for deployment: forecasting systems must operate across the full seasonal cycle without knowing in advance which regime the next window will occupy. A model that is near-best in every regime, and best where errors are largest, offers higher expected value than phase-specific specialists that may be substantially weaker elsewhere. This is especially important in ILI forecasting because the peak phase is both the hardest and the most operationally consequential. DSA reduces peak-phase MSE to 6.738, compared with 11.953 for TaTS, 14.405 for iTransformer, and 28.591 for GPT4MTS, meaning that the largest public-health-relevant errors are controlled most effectively by DSA. The preliminary no-retuning Health\_Africa evaluation is also consistent with transferability of the learned fusion strategy.
	
	The ablation and faithfulness analyses clarify what these performance differences mean. DSA retains a substantial advantage when BioLinkBERT is frozen and remains competitive when the text encoder is replaced by general-domain alternatives, supporting the view that the cross-modal interaction mechanism is a major contributor rather than a consequence of a particular language model. A single-seed ablation of the attention mechanism itself further indicates that the bidirectional design contributes in its own right: both single-direction variants underperform full DSA (Section 5.4), though this specific comparison has not yet been validated across multiple seeds. Targeted perturbations further show that highly attended historical information is functionally informative under the tested perturbations, especially for the Text$\rightarrow$Numerical pathway. We read this deliberately as a non-causal, functional-importance result: attention in that direction skews toward recent weeks, which is broadly consistent with weekly surveillance narrative being written to describe the current state rather than a retrospective account, though the present dataset does not let us fully rule out that recency alone, rather than the attention mechanism specifically, accounts for part of the effect.
	
	The results therefore support an important conclusion for multimodal epidemic forecasting: the value of unstructured surveillance text depends not only on whether it is included, but on how its information is conditioned on the numerical state of the system. Future work should test how far this generalizes: to additional diseases and geographies within epidemiological surveillance, and more broadly to other domains where a numerical time series is routinely paired with free text, which would likely require a domain-appropriate text encoder in place of BioLinkBERT. A second direction is extending the forecast horizon well beyond the 12 weeks demonstrated here without letting error grow with it; this will likely require a richer decoding mechanism than the single pooled representation used in this study, since our own attempts to push past $H=12$ with the current decoder did not hold up (Section 5.7). Such extensions would test whether the cross-modal conditioning principle behind DSA is a property of this dataset or a more general one.

				\appendix
				\section*{Supplementary Material}
				\addcontentsline{toc}{section}{Supplementary Material}

				\section{Training Stability and Reproducibility}
				\label{app:training_stability}
				
				The final DSA configuration was selected only after a reproducibility investigation aimed at distinguishing architectural effects from ordinary stochastic variation. This investigation was motivated by substantial seed-to-seed variation observed during early development. Rather than treating the lowest individual run as representative, we examined the effect of training-control choices before fixing the configuration used for the ten-seed evaluation.
				
				\subsection{Sources of run-to-run variation}
				
				DSA is trained with AdamW, a learning rate of $10^{-5}$, dropout of $0.15$, cosine learning-rate scheduling, a 10-epoch warmup, automatic mixed precision, and gradient accumulation. The micro-batch size is four and the final configuration uses four accumulation steps, giving an effective batch size of 16. The reproducibility investigation identified an interaction between the learning-rate schedule and the number of gradient-accumulation steps as an important source of early seed-to-seed variation. This observation motivated fixing the accumulation configuration before the final ten-seed comparison rather than selecting a seed-specific training setup.
				
				Importantly, the purpose of this investigation was not to optimize each seed independently. Once the final configuration was selected, the same training protocol was applied to all seeds in the main comparison. This prevents the reported ten-seed distribution from becoming a collection of independently tuned runs.
				
				\subsection{Final reproducibility configuration}
				
				The final configuration used for the principal Health\_US experiments is summarized in Table~\ref{tab:training_config}. Values that were not varied or independently reported in the manuscript are intentionally omitted rather than inferred.
				\begin{table}[t]
					\caption{Training configuration used for the principal DSA evaluation.}
					\label{tab:training_config}
					\small
					\begin{tabular*}{\tblwidth}{@{} L C @{}}
						\toprule
						\textbf{Setting} & \textbf{Final configuration} \\
						\midrule
						Lookback window & 36 weeks \\
						Forecast horizon & 12 weeks \\
						Numerical input dimension & 14 \\
						Numerical encoder & 4-layer Transformer \\
						Numerical hidden dimension & 512 \\
						Text encoder & BioLinkBERT-base \\
						Text maximum length & 256 tokens \\
						BioLinkBERT fine-tuning & Last 8 layers unfrozen \\
						Text pooling & Masked mean pooling \\
						Inter-week text attention & 1 layer \\
						Fusion dimension & 1024 \\
						Causal temporal encoder & 2-layer Transformer \\
						Forecast decoder & 2-layer MLP \\
						Optimizer & AdamW \\
						Learning rate & $1\times10^{-5}$ \\
						Dropout & $0.15$ \\
						LR schedule & Cosine \\
						Warmup & 10 epochs \\
						Micro-batch size & 4 \\
						Gradient accumulation & 4 steps \\
						Effective batch size & 16 \\
						Mixed precision & AMP \\
						Primary training loss & MSE \\
						\bottomrule
					\end{tabular*}
				\end{table}
				
				\subsection{Random-seed protocol}
				
				The principal comparison uses ten independent random seeds(Namely, 0,1,2,3,5,7,10,13,21,42).
				For each seed, the same chronological train/validation/test split, preprocessing procedure, architecture, optimizer configuration, scheduler configuration, and evaluation procedure are retained. The purpose of the seed sweep is therefore to estimate the stability of the observed architectural difference rather than to search for a favorable initialization.

				\subsection{Why the seed distribution matters}
				
				The ten-seed experiment is particularly important because a single DSA run can give an incomplete picture of its expected performance. Accordingly, the principal result is not based on the best seed. The manuscript reports the median test MSE, the mean and standard deviation of horizon-averaged error, tail-risk statistics, and the directional consistency of DSA relative to the three core baselines. Across the ten seeds, DSA wins against iTransformer, TaTS, and GPT4MTS in all ten matched comparisons. This directional consistency is complementary to the effect-size analysis and avoids interpreting one unusually favorable initialization as evidence of a general architectural advantage.
				
				\section{Additional Baseline and Ablation Analyses}
				\label{app:additional_ablations}
				
				\subsection{Single-seed screening protocol}
				\label{app:screening}
				
				The complete baseline space considered during model development was broader than the four models used in the ten-seed statistical comparison. Because evaluating every candidate across ten seeds would substantially increase computational cost, we used a two-stage evaluation protocol.
				
				In the first stage, candidate configurations were evaluated under a fixed seed. The purpose of this stage was to identify a strong representative of each model family, rather than to make final claims from a single run. The candidate space included:
				
				\begin{enumerate}
					\item unimodal numerical forecasters, including DLinear, PatchTST, and iTransformer;
					\item naive multimodal fusion variants combining BERT-family text representations with numerical forecasting backbones;
					\item TaTS variants using different numerical backbones;
					\item GPT4MTS as the LLM-based forecasting representative; and
					\item the proposed DSA architecture.
				\end{enumerate}
				
				\begin{table}[width=\linewidth,cols=4,pos=h]
					\caption{Single-seed screening results on the US test set. Lower values indicate better performance.}
					\label{tab:us_single_seed}
					\small
					\begin{tabular*}{\tblwidth}{@{} L C C C @{}}
						\toprule
						\textbf{Model} &
						\textbf{MSE} &
						\textbf{MAE} &
						\textbf{RMSE} \\
						\midrule
						
						DSA (Multimodal) &
						$\mathbf{1.0323}$ &
						$\mathbf{0.7099}$ &
						$\mathbf{1.0160}$ \\
						
						DSA (Unimodal) &
						$1.0852$ &
						$0.7397$ &
						$1.0417$ \\
						
						iTransformer (Unimodal) &
						$1.1111$ &
						$0.7362$ &
						$1.0541$ \\
						
						PatchTST (Unimodal) &
						$1.3181$ &
						$0.8026$ &
						$1.1481$ \\
						
						DLinear (Unimodal) &
						$1.6642$ &
						$0.8216$ &
						$1.2900$ \\
						
						iTransformer (Naive MM Fusion) &
						$2.4165$ &
						$1.1205$ &
						$1.5545$ \\
						
						PatchTST (Naive MM Fusion) &
						$2.9825$ &
						$1.4020$ &
						$1.7270$ \\
						
						DLinear (Naive MM Fusion) &
						$2.2620$ &
						$1.1560$ &
						$1.5040$ \\
						
						TaTS (iTransformer) &
						$1.2772$ &
						$0.7625$ &
						$1.1301$ \\
						
						TaTS (PatchTST) &
						$2.5416$ &
						$1.1697$ &
						$1.5942$ \\
						
						TaTS (DLinear) &
						$2.1114$ &
						$0.9793$ &
						$1.4531$ \\
						
						GPT4MTS &
						$1.3400$ &
						$0.8128$ &
						$1.1576$ \\
						
						\bottomrule
					\end{tabular*}
				\end{table}
				
				The screening stage selected DSA, iTransformer, TaTS with an iTransformer backbone, and GPT4MTS as the four configurations for the principal ten-seed evaluation. We deliberately distinguish this screening procedure from the final comparison. The screening identifies which representatives justify the expensive multi-seed analysis; it is not itself treated as a ten-seed statistical experiment.
				
				\subsection{Architecture sensitivity}
				\label{app:architecture}
				
				Table~\ref{tab:supp_architecture} summarizes the principal architecture-search results reported in the manuscript.
				\begin{table*}[t]
					\caption{Architecture-search results}
					\label{tab:supp_architecture}
					\small
					\begin{tabular*}{\textwidth}{@{} L C L @{}}
						\toprule
						\textbf{Modification} &
						\textbf{Test MSE} &
						\textbf{Interpretation} \\
						\midrule
						Remove causal mask &
						$1.7092$ &
						Causal masking is load-bearing \\
						
						Add RevIN &
						$1.7992$ &
						Degrades performance \\
						
						CLS pooling &
						$1.2825$ &
						Mean pooling is preferable \\
						
						iTransformer-style numerical encoder &
						$1.8320$ &
						Rejected \\
						
						Numerical depth $4\rightarrow6$ &
						$1.1940$ &
						No improvement; overfitting \\
						
						BERT learning rate $1\times10^{-6}$ &
						Degraded &
						Full learning rate retained \\
						
						CMA: Text$\rightarrow$Numerical only &
						$1.0971$ &
						Worse than bidirectional (Appendix~\ref{app:directional_cma}) \\
						
						CMA: Numerical$\rightarrow$Text only &
						$1.1508$ &
						Worse than bidirectional (Appendix~\ref{app:directional_cma}) \\
						\bottomrule
					\end{tabular*}
				\end{table*}
				The strongest qualitative result in this search (Table~\ref{tab:supp_architecture}) is the effect of the causal mask. Removing the mask increases the reported test MSE from approximately 1.09 for the corresponding DSA configuration to 1.7092. The pooling experiment provides a second useful constraint. Replacing masked mean pooling with CLS-token pooling increases the reported test MSE to 1.2825. The final model therefore uses masked mean pooling over the BioLinkBERT token representations. The deeper numerical encoder also failed to improve performance: increasing the numerical Transformer depth from four to six layers gives a test MSE of 1.1940. This result was interpreted as evidence that increasing numerical-model capacity was not beneficial at the available dataset size. The two CMA-directionality rows in Table~\ref{tab:supp_architecture} are discussed in full in Appendix~\ref{app:directional_cma}.
				
				\subsection{Directional Cross-Modal Attention}
				\label{app:directional_cma}
				
				DSA's fusion mechanism is deliberately bidirectional: the numerical stream queries the text stream (Eq.~6a) and the text stream queries the numerical stream (Eq.~6b), and both outputs are concatenated before the fusion projection (Eq.~7). To test whether this bidirectional design earns its added complexity over a single attention direction, we constructed two ablated variants that each retain only one attention branch. Each unidirectional variant uses a properly sized fusion projection matched to a single branch's output dimension, rather than bidirectional's projection with one branch's input dropped, so the comparison isolates directionality rather than parameter count.
				
				Table~\ref{tab:supp_directional_cma} reports test MSE, MAE, and RMSE for full bidirectional DSA and both unidirectional variants at a single seed (SEED=42), matching the fixed-seed architecture-search protocol described in Section 4.3. This ablation has not been repeated across multiple seeds, so the comparison below should be read as a single-run directional result rather than as a statistically validated claim; no significance test is reported for it.
				
				\begin{table}[t]
					\caption{Directional CMA ablation}
					\label{tab:supp_directional_cma}
					\small
					\begin{tabular*}{\tblwidth}{@{} L C C C @{}}
						\toprule
						\textbf{CMA configuration} &
						\textbf{Test MSE} &
						\textbf{MAE} &
						\textbf{RMSE} \\
						\midrule
						\textbf{Bidirectional (full DSA)} &
						\textbf{$\mathbf{1.0367}$} &
						\textbf{$\mathbf{0.7131}$} &
						\textbf{$\mathbf{1.0182}$} \\
						
						Text$\rightarrow$Numerical only &
						$1.0971$ &
						$0.7534$ &
						$1.0474$ \\
						
						Numerical$\rightarrow$Text only &
						$1.1508$ &
						$0.7821$ &
						$1.0728$ \\
						\bottomrule
					\end{tabular*}
				\end{table}
				
				Both unidirectional variants underperform full bidirectional DSA on all three metrics. The ordering also mirrors the faithfulness asymmetry reported in Section 5.6 and Supplementary Appendix~\ref{app:faithfulness}: the Text$\rightarrow$Numerical-only variant, which keeps the direction found to be more informative under perturbation, degrades less than the Numerical$\rightarrow$Text-only variant, which keeps the less informative direction. At this seed, dropping either direction costs accuracy, so bidirectionality is not redundant; whether this pattern holds across seeds has not yet been tested.
				
				\subsection{Frozen text encoder}
				\label{app:frozen_encoder}
				
				To separate the contribution of language-model adaptation from the contribution of cross-modal fusion, we retrained DSA with the BioLinkBERT parameters frozen. The remainder of the architecture and forecasting objective was retained.
				
				\begin{figure*}
					\centering
					\includegraphics[width=\textwidth]{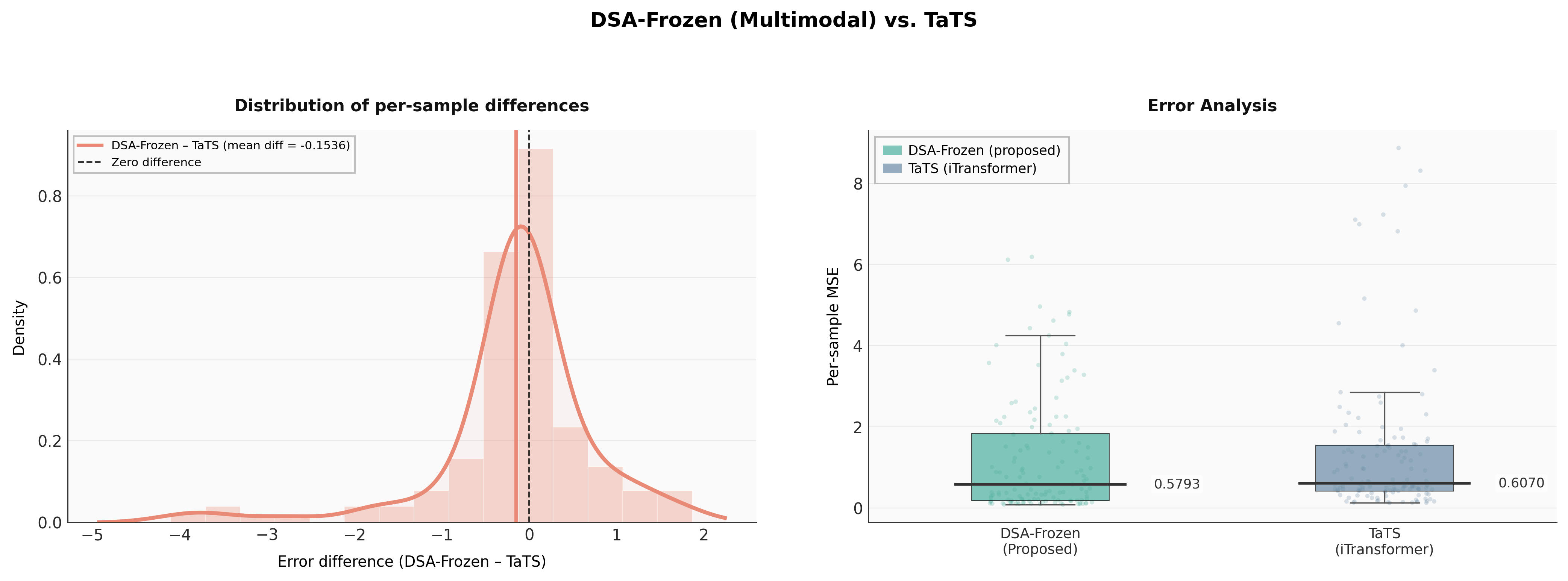}
					\caption{Comparing DSA-Frozen (multimodal) with TaTS (iTransformer), where both models employ frozen encoders. (Left) Distribution of per-sample error differences (DSA-Frozen minus TaTS), showing a mean difference of -0.1536. (Right) Box plot comparison of per-sample MSE, demonstrating that DSA-Frozen achieves a slightly lower average error (0.5793) compared to TaTS (0.6070).}
					\label{FIG:12}
				\end{figure*}
				
				\begin{figure*}
					\centering
					\includegraphics[width=\textwidth]{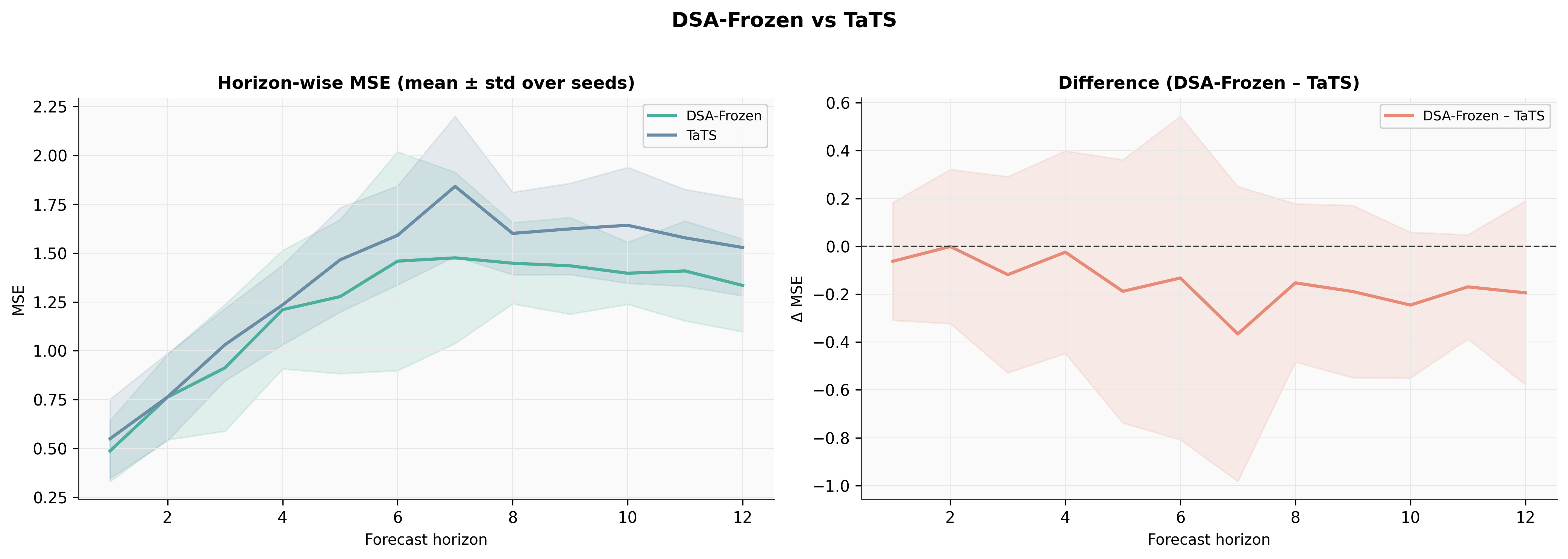}
					\caption{Horizon-wise comparison of DSA-Frozen versus TaTS. The left panel shows DSA-Frozen achieves lower and more stable MSE across all horizons, with the right panel confirming consistent negative error differences, indicating its superiority over TaTS, particularly at longer forecast steps.}
					\label{FIG:13}
				\end{figure*}
				
				\begin{figure}
					\centering
					\includegraphics[width=\columnwidth]{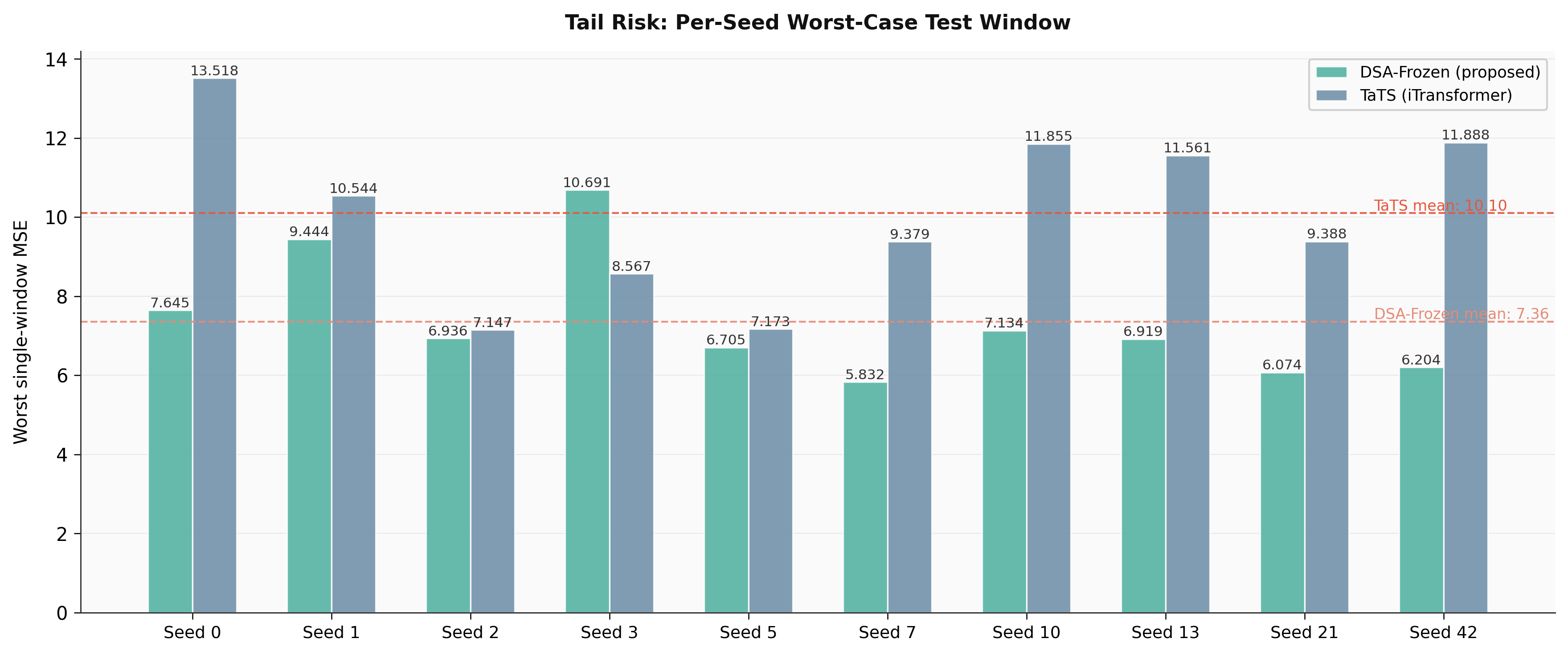}
					\caption{Tail risk evaluation of DSA-Frozen vs. TaTS across random seeds. DSA-Frozen consistently yields lower worst-case single-window MSE, achieving a notably lower overall mean (7.36) compared to TaTS (10.10).}
					\label{FIG:14}
				\end{figure}
				The frozen-text variant achieves a ten-seed mean MSE of $1.217 \pm 0.194$ (Figures~\ref{FIG:12}--\ref{FIG:14}),
				compared with $1.120 \pm 0.096$ for the fully specified DSA configuration and $1.371 \pm 0.177$ for TaTS with an iTransformer backbone.
				This comparison does not imply that fine-tuning is unimportant: freezing BioLinkBERT removes task-specific adaptation and produces a measurable loss relative to full DSA (1.217 vs.\ 1.120), so fine-tuning does contribute part of the gain. What the comparison shows is that this contribution is the smaller of the two: even without it, the frozen variant still beats the strongest fusion baseline by a wider margin (1.217 vs.\ 1.371) than fine-tuning itself adds (1.217 vs.\ 1.120). The majority of DSA's advantage over TaTS is therefore attributable to the cross-modal attention mechanism itself, with fine-tuning contributing a smaller, additional improvement on top of it.
				
				\subsection{Text-encoder sensitivity}
				\label{app:text_encoder}
				
				To determine whether the observed performance depends specifically on BioLinkBERT, the DSA pipeline was repeated with BERT-base-uncased and MiniLM-L6-v2. These experiments used five seeds rather than the ten seeds of the principal comparison. This subsection reports the full numeric detail behind the encoder-ablation summary and Figure~\ref{FIG:9} in the main text (Section 5.4).
				
				\begin{figure}
					\centering
					\includegraphics[width=\columnwidth]{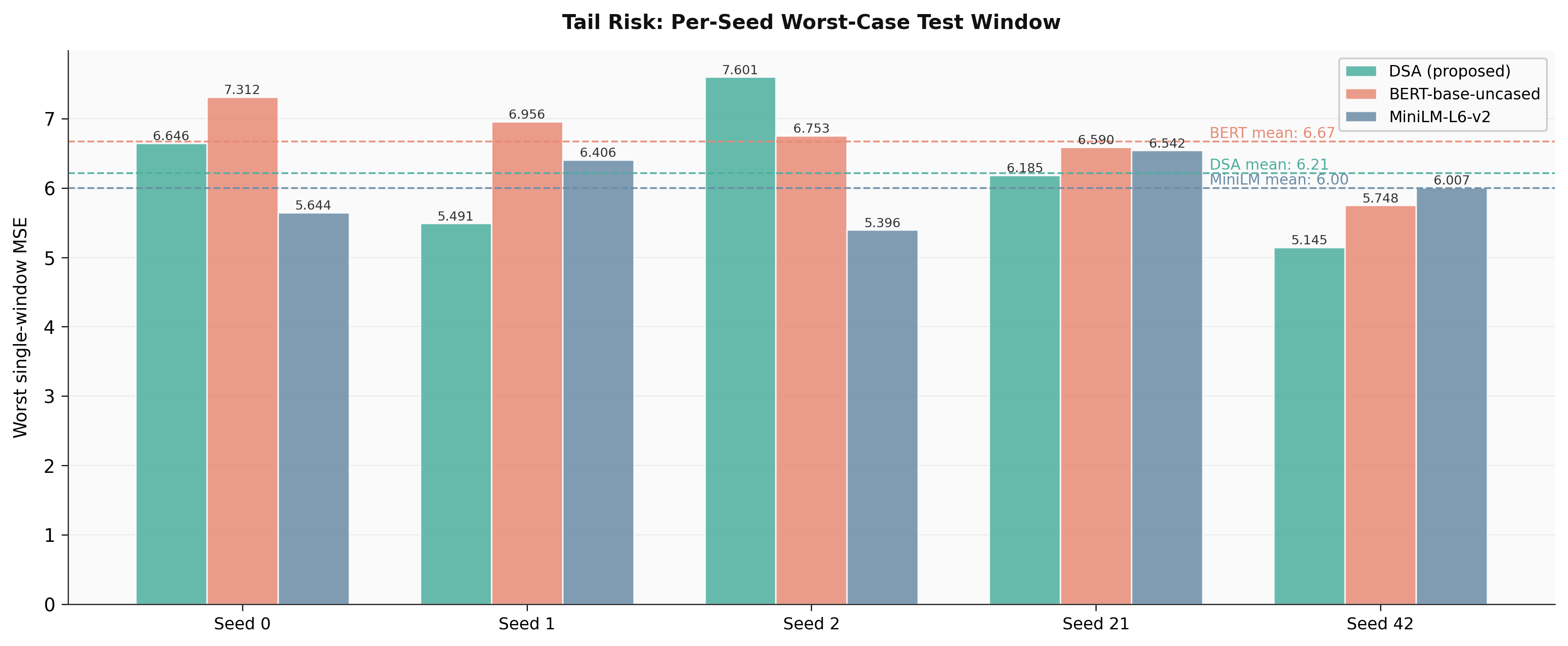}
					\caption{Tail risk assessment comparing the proposed DSA with BERT-base-uncased and MiniLM-L6-v2 across random seeds. MiniLM-L6-v2 achieves the lowest mean worst-case MSE (6.00), closely followed by DSA (6.21), whereas BERT exhibits the highest tail risk (6.67).}
					\label{FIG:15}
				\end{figure}
				
				\begin{table*}[t]
					\caption{Five-seed text-encoder sensitivity analysis.}
					\label{tab:supp_text_encoders}
					\small
					\begin{tabular*}{\tblwidth}{@{} L C C C @{}}
						\toprule
						\textbf{Text Encoder} &
						\textbf{Median MSE} &
						\textbf{Max MSE (mean $\pm$ s.d.)} &
						\textbf{P90 MSE (mean $\pm$ s.d.)} \\
						\midrule
						BioLinkBERT &
						\textbf{$\mathbf{0.367}$} &
						$6.2136 \pm 0.8690$ &
						\textbf{$\mathbf{2.9928} \pm \mathbf{0.1062}$} \\
						
						BERT-base-uncased &
						$0.445$ &
						$6.6717 \pm 0.5211$ &
						$3.0148 \pm 0.1882$ \\
						
						MiniLM-L6-v2 &
						$0.505$ &
						$\mathbf{5.9991} \pm \mathbf{0.4359}$ &
						$3.3307 \pm 0.2492$ \\
						\bottomrule
					\end{tabular*}
				\end{table*}
				
				The results (Table~\ref{tab:supp_text_encoders}; Figure~\ref{FIG:15}) indicate that BioLinkBERT remains the strongest of the three encoders on median MSE and P90 MSE. However, replacing it with either general-domain encoder does not eliminate the advantage of the DSA fusion architecture. The MiniLM configuration also produces the lowest mean worst-case MSE among the three (Figure~\ref{FIG:15}), illustrating that encoder choice does not induce a uniform ordering across all metrics.
				
				\subsection{Health\_Africa supplementary analysis}
				\label{app:africa_tuning}
				\label{app:africa_error}
				
				The external Health\_Africa experiment was designed as a no-retuning evaluation. In addition to the primary comparison, we conducted a limited supplementary investigation of the TaTS text-to-numerical dimension mapping to have a fair comparative baseline. The mapping dimension was increased from 12 to 32 for the three TaTS numerical backbones to investigate even larger capacities. The change did not produce a consistent improvement and was therefore not adopted as a revised TaTS configuration for the external comparison. This result does not establish that the chosen dimension is globally optimal for TaTS; it only indicates that the tested increase did not resolve the observed cross-geography weakness.
				Because Health\_Africa is a single external dataset and the reported evaluation does not follow a full 10 seed run, these results are treated as preliminary evidence of transfer rather than as a fully grounded external validation claim.			
				The primary Health\_Africa result is that multimodal DSA has the lowest reported test MSE, MAE, and RMSE among the compared configurations:
				
				\begin{table*}[t]
					\caption{Health\_Africa external evaluation. Lower values indicate better forecasting performance.}
					\label{tab:supp_africa}
					\small
					\begin{tabular*}{\tblwidth}{@{} L C C C @{}}
						\toprule
						\textbf{Model} &
						\textbf{MSE} &
						\textbf{MAE} &
						\textbf{RMSE} \\
						\midrule
						DSA (proposed) &
						\textbf{$\mathbf{0.0040}$} &
						\textbf{$\mathbf{0.0543}$} &
						\textbf{$\mathbf{0.0632}$} \\
						
						GPT4MTS &
						$0.0050$ &
						$0.0589$ &
						$0.0704$ \\
						
						DSA (Unimodal) &
						$0.0053$ &
						$0.0556$ &
						$0.0725$ \\
						
						DLinear &
						$0.0086$ &
						$0.0817$ &
						$0.0930$ \\
						
						TaTS + DLinear &
						$0.0140$ &
						$0.1116$ &
						$0.1183$ \\
						
						PatchTST &
						$0.0175$ &
						$0.1069$ &
						$0.1321$ \\
						
						iTransformer &
						$0.0254$ &
						$0.1375$ &
						$0.1592$ \\
						
						TaTS + PatchTST &
						$0.0279$ &
						$0.1329$ &
						$0.1670$ \\
						
						TaTS + iTransformer &
						$0.0407$ &
						$0.1709$ &
						$0.2018$ \\
						\bottomrule
					\end{tabular*}
				\end{table*}

				\section{Additional Statistical Analyses}
				\label{app:statistics}
				
				\subsection{Per-seed directional consistency}
				
				The ten-seed evaluation produces a favorable direction for DSA in all ten matched comparisons against each of the three core baselines. The exact two-sided sign-test probability for observing ten wins out of ten under a null probability of one half is
				\[
				p =
				2\left(\frac{1}{2}\right)^{10}
				=
				0.001953125.
				\]
				
				Thus, the directional consistency is unlikely under a null model in which DSA and the corresponding baseline are equally likely to win each seed. This test is intentionally complementary to the DM analysis: it tests the direction of the paired seed-level outcome rather than the magnitude or temporal dependence of the forecast-error differential.
				
				\subsection{Diebold--Mariano analysis}
				
				The main DM analysis operates on paired forecast-error differentials:
				\[
				d_\tau=e_{DSA,\tau}^{2}-e_{baseline,\tau}^{2}.
				\]
				A negative mean differential favors DSA. Because the test windows overlap in time and each forecast contains multiple future horizons, the loss differential can exhibit serial dependence. We therefore use the Newey--West long-run variance estimator with truncation lag:
				\[
				q=H-1=11.
				\]
				This correction is important for interpretation. A naive lag-0 variance estimate can produce smaller standard errors when the loss differential is positively autocorrelated. The corrected analysis is consequently more conservative and should be preferred when reporting inferential claims. We combine effect sizes, directional consistency, median performance, tail-risk, bootstrap rank probabilities, MCS inclusion, and DM testing into a single coherent picture rather than relying on any one statistic in isolation.
				
\begin{figure}
					\centering
					\includegraphics[width=\columnwidth]{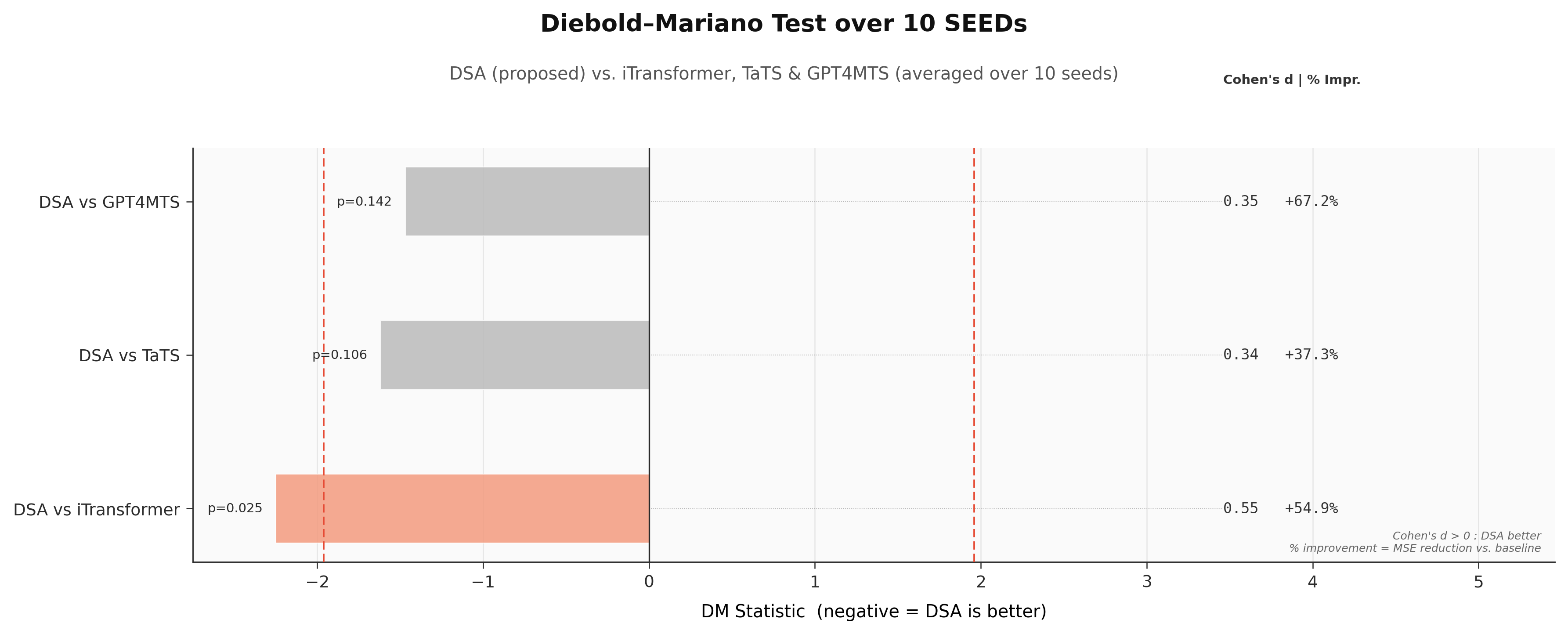}
					\caption{Diebold--Mariano test statistics (Newey--West corrected, ten-seed mean) for DSA against iTransformer, TaTS, and GPT4MTS. Negative values favor DSA; dashed lines mark the conventional $\pm1.96$ critical value at $\alpha=0.05$. Cohen's $d$ and \% improvement are reproduced from Table~\ref{tab:effect_size} for reference.}
					\label{FIG:16}
				\end{figure}

				Table~\ref{tab:supp_dm} and Figure~\ref{FIG:16} report the Newey--West-corrected DM statistic and two-sided $p$-value for each of the three paired comparisons, averaged over the ten seeds.
				
				\begin{table}[t]
					\caption{Diebold--Mariano test statistics (Newey--West corrected), ten-seed mean.}
					\label{tab:supp_dm}
					\small
					\begin{tabular*}{\tblwidth}{@{} L C C @{}}
						\toprule
						\textbf{DSA vs} &
						\textbf{DM statistic} &
						\textbf{$p$-value} \\
						\midrule
						iTransformer &
						$-2.241$ &
						$0.025$ \\
						
						TaTS &
						$-1.616$ &
						$0.106$ \\
						
						GPT4MTS &
						$-1.468$ &
						$0.142$ \\
						\bottomrule
					\end{tabular*}
				\end{table}
				
				Only the DSA-versus-iTransformer comparison crosses the conventional two-sided $\alpha=0.05$ threshold at the level of individual paired differentials; the other two do not. This is the expected signature of a conservative, autocorrelation-robust variance correction applied to a ten-seed comparison rather than evidence against DSA's advantage: the DM statistic tests only the magnitude and precision of the mean loss differential, and it is not the only --- or even the primary --- piece of evidence we rely on. All three DM statistics are negative, consistent with the 10/10 directional win rate reported above; the effect sizes in Table~\ref{tab:effect_size} are medium-to-large by conventional benchmarks ($d=0.34$--$0.56$); and DSA is ranked first in every bootstrap draw (Table~\ref{tab:rank_probability}) and remains in the Model Confidence Set at every tested threshold. Read together, these independent statistical lenses tell a consistent story even where the per-seed DM test alone is inconclusive at conventional significance for two of the three baselines.
				
				For each baseline, the paired Cohen's $d$ values reported in the main manuscript (Table~\ref{tab:effect_size}) are calculated on matched differences rather than treating the ten model runs as independent samples. This preserves the paired nature of the experiment. Furthermore, bootstrap ranking was performed over the ten-seed evaluation and considered all four core models jointly. The resulting rank probabilities are visualized in Figure~\ref{FIG:3}.
				
				It is worth mentioning that the rank analysis answers a different question from the DM test: rather than asking whether a particular pair differs significantly, it asks how frequently each model occupies each rank under resampling of the observed seed-level performance. DSA occupying rank one in every reported bootstrap draw provides a strong descriptive indication of ranking stability. It should not, however, be interpreted as equivalent to a formal probability that DSA is the universally optimal forecasting architecture.
				
				Coming to the next step, Model Confidence Set (MCS) analysis is used to determine whether models can be statistically eliminated from the set of models that are indistinguishable from the best model at a specified confidence level.
				
				The main analysis retains DSA as the only model in the tested MCS across the reported confidence levels. Phase-specific analyses provide a more nuanced result: DSA remains in the MCS at all tested thresholds in the off-season, rising, and peak phases, while its membership becomes less stable in the declining phase. This distinction is important. MCS membership is not a statement that the retained model is statistically superior to every excluded model under every possible loss function. It indicates that, under the specified loss differential and confidence threshold, the model cannot be eliminated by the MCS procedure.
				
				\section{Tail-Risk and Horizon-Wise Supplementary Results}
				\label{app:tail_horizon}
				
				\subsection{Worst-window behavior}
				
				The ten-seed analysis provides a complementary view of robustness by examining the maximum single-window MSE encountered within each seed. Table~\ref{tab:worst_window_by_seed} reports these values for DSA and the three baselines. The largest value observed across the ten seeds is 7.703 for DSA, compared with 15.643 for iTransformer, 13.748 for TaTS, and 20.407 for GPT4MTS. Thus, DSA exhibits a substantially lower worst-case single-window error across random initializations.
				
				These values should be distinguished from the mean maximum-window errors reported in Table~\ref{tab:overall_metrics}. The latter are computed from the underlying, unrounded evaluation results and summarize the average worst-window behavior across seeds. They are 6.4036 for DSA, 10.7543 for iTransformer, 10.1020 for TaTS, and 14.5867 for GPT4MTS. The per-seed results in Table~\ref{tab:worst_window_by_seed} provide the corresponding distribution across random seeds, while the aggregate values in Table~\ref{tab:overall_metrics} provide its mean.
				
				\begin{table}[t]
					\centering
					\caption{Maximum single-window MSE observed within the test set for each random seed. Lower values indicate better worst-window robustness.}
					\label{tab:worst_window_by_seed}
					\begin{tabular}{lrrrr}
						\toprule
						Seed & DSA & iTransformer & TaTS & GPT4MTS \\
						\midrule
						0  & $6.646$ & $6.308$ & $13.518$ & $15.920$ \\
						1  & $5.491$ & $11.840$ & $10.544$ & $16.377$ \\
						2  & $7.601$ & $8.138$ & $7.147$ & $11.315$ \\
						3  & $5.989$ & $12.042$ & $8.567$ & $15.852$ \\
						5  & $6.957$ & $14.297$ & $7.173$ & $11.279$ \\
						7  & $6.191$ & $12.399$ & $9.379$ & $13.454$ \\
						10 & $7.703$ & $8.637$ & $11.855$ & $11.883$ \\
						13 & $6.128$ & $15.643$ & $11.561$ & $20.407$ \\
						21 & $6.185$ & $10.800$ & $9.388$ & $19.318$ \\
						42 & $5.145$ & $7.402$ & $11.888$ & $10.061$ \\
						\midrule
						Maximum & $\mathbf{7.703}$ & $15.643$ & $13.518$ & $20.407$ \\
						\bottomrule
					\end{tabular}
				\end{table}
				
				\subsection{P90 error}
				
				P90 MSE is the 90th percentile of the per-window squared-error distribution for a given model, computed by pooling all per-window errors across the ten seeds and reading off the value below which 90\% of them fall. It sits between the median and the single worst-case (Max) error reported in Table~\ref{tab:overall_metrics}: unlike the median, it is sensitive to the upper tail of the error distribution, and unlike the Max column, it is not driven by a single extreme window, so it captures how badly a model tends to perform on its harder test windows without being dominated by one outlier. DSA has the lowest reported P90 error of the four core models (Table~\ref{tab:overall_metrics}), consistent with its lowest horizon-averaged error over the same comparison.
				
				\subsection{Horizon-wise behavior}
				
				The main horizon-wise analysis evaluates each of the 12 forecast positions independently. The reported qualitative pattern is:
				
				\begin{itemize}
					\item DSA and iTransformer are close at horizon 1, after which the DSA advantage increases and reaches its largest reported gap around horizon 10.
					\item The largest reported DSA advantage over TaTS occurs around horizon 7.
					\item The largest reported DSA advantage over GPT4MTS occurs around horizon 11.
				\end{itemize}
				
				No forecast horizon in the reported analysis has a baseline with a lower average MSE than DSA. The error margin is growing over the horizon rather than as a uniformly large improvement at every horizon.
				
				\section{Phase-Stratified Analysis}
				\label{app:phase}
				
				The test windows are categorized into four epidemic regimes: off-season, rising, peak, and declining. The phase analysis is intended to test whether aggregate performance is driven by one particular part of the seasonal cycle.
				
				\begin{table*}[t]
					\caption{Supplementary phase-stratified MSE values.}
					\label{tab:supp_phase}
					\small
					\begin{tabular*}{\tblwidth}{@{} L C C C C @{}}
						\toprule
						\textbf{Phase} &
						\textbf{DSA} &
						\textbf{iTransformer} &
						\textbf{TaTS} &
						\textbf{GPT4MTS} \\
						\midrule
						Off-season &
						$1.917 \pm 0.679$ &
						$5.605 \pm 4.616$ &
						$4.019 \pm 2.301$ &
						$\mathbf{1.778} \pm \mathbf{1.245}$ \\
						
						Rising &
						$0.624 \pm 0.277$ &
						$2.787 \pm 3.053$ &
						$\mathbf{0.535} \pm \mathbf{0.334}$ &
						$1.747 \pm 1.086$ \\
						
						Peak &
						$\mathbf{6.738} \pm \mathbf{3.087}$ &
						$14.405 \pm 7.926$ &
						$11.953 \pm 5.067$ &
						$28.591 \pm 12.333$ \\
						
						Declining &
						$2.260 \pm 1.263$ &
						$12.170 \pm 12.146$ &
						$2.957 \pm 2.394$ &
						$\mathbf{0.967} \pm \mathbf{0.529}$ \\
						\bottomrule
					\end{tabular*}
				\end{table*}
				
				The phase results demonstrate why aggregate metrics should not be interpreted as evidence of uniform dominance. DSA is best in the off-season, rising, and peak phases according to the reported mean MSE, but GPT4MTS is substantially better during the declining phase and marginally better during the off-season. TaTS is better than DSA during the rising phase.
				
				The peak phase is particularly important for interpreting the aggregate result because it has the largest reported errors for all four models. DSA reduces the peak-phase mean MSE to 6.738, compared with 11.953 for TaTS, 14.405 for iTransformer, and 28.591 for GPT4MTS. For the rising phase, the reported Cohen's $d$ for DSA relative to TaTS is $-0.19$, with TaTS's mean MSE approximately 16.8\% lower. For the declining phase, GPT4MTS has a substantially lower error than DSA, with the reported Cohen's $d=-1.13$. 
				
				\section{CMA Faithfulness Analysis}
				\label{app:faithfulness}
				
				\subsection{Perturbation protocol}
				
				The purpose of the perturbation analysis is to test whether the attention weights learned by CMA are functionally informative. Attention weights alone do not establish that the attended information is used by the forecasting function. We therefore compare the effect of masking weeks selected according to their attention values. For each test window, weeks are divided into three groups:
				
				\begin{enumerate}
					\item \textbf{Top-$k$:} the $k$ weeks receiving the largest attention weights;
					\item \textbf{Random-$k$:} $k$ randomly selected weeks; and
					\item \textbf{Bottom-$k$:} the $k$ weeks receiving the smallest attention weights.
				\end{enumerate}
				
				The forecast is recomputed after masking the selected information and the change in MSE relative to the unperturbed prediction is recorded. The analysis is performed separately for the two CMA pathways. The primary reported masking budget is $k=3$, corresponding to 3 of the 36 historical weeks, or approximately 8.3\% of the lookback window. Additional budgets from $k=1$ through $k=12$ are also evaluated.
				
				\subsection{Three-week masking results}
				
				The two directions exhibit the same qualitative ordering:
				\[
				\mathrm{Top\text{-}}k >
				\mathrm{Random\text{-}}k >
				\mathrm{Bottom\text{-}}k.
				\]
				However, the magnitude of the effect is markedly asymmetric (Table~\ref{tab:supp_perturbation}). For $k=3$, masking the top-attended weeks in the Text$\rightarrow$Numerical pathway increases MSE by 1.4436, whereas the corresponding Numerical$\rightarrow$Text increase is 0.0352. The difference is approximately
				\[
				\frac{1.4436}{0.0352}\approx 41.0,
				\]
				so the Text$\rightarrow$Numerical perturbation effect is roughly $41\times$ larger for the reported $k=3$ experiment. This asymmetry is plausible given what each pathway is masking: removing three weeks of numerical history is a small, roughly linear perturbation to a 14-dimensional signal the model has 36 weeks of, whereas removing the corresponding weeks of text removes the specific, concentrated narrative content those weeks carry, with no equally informative numerical substitute available to the Numerical$\rightarrow$Text pathway. The gap is also consistent with the recency correlation discussed below: because Text$\rightarrow$Numerical attention concentrates on recent weeks (Section~\ref{app:faithfulness_limitations}, below), and the present dataset's test windows are not long or varied enough to fully disentangle recency from content, part of the $41\times$ gap likely reflects how much the model leans on recent surveillance text specifically, rather than indicating that the two pathways are mismatched in reliability.
				
				\begin{table}[t]
					\caption{Change in test MSE following targeted perturbation of the three most, randomly, or least attended weeks.}
					\label{tab:supp_perturbation}
					\small
					\begin{tabular*}{\tblwidth}{@{} L C C C C C @{}}
						\toprule
						\textbf{Direction} &
						\textbf{Top-3} &
						\textbf{Rand-3} &
						\textbf{Bottom-3} &
						\textbf{$p$} &
						\textbf{$d$} \\
						\midrule
						Num $\rightarrow$ Text &
						$0.0352$ &
						$0.0199$ &
						$0.0171$ &
						$0.0126$ &
						$0.243$ \\
						
						Text $\rightarrow$ Num &
						$1.4436$ &
						$0.1665$ &
						$0.0025$ &
						$2.06\times10^{-7}$ &
						$0.411$ \\
						\bottomrule
					\end{tabular*}
				\end{table}

				\subsection{Masking-budget sensitivity}
				
				The same qualitative ordering is reported for masking budgets from $k=1$ to $k=12$, corresponding to approximately $2.8\%$ to $33.3\%$ of the 36-week lookback window. The reported effect becomes larger as more historical weeks are removed. This result supports the interpretation that the attention rankings contain useful information about which historical weeks the model relies upon. It does not demonstrate that the selected weeks are individually necessary, because masking several weeks simultaneously can produce nonlinear interactions and redundancy among information sources.
				
				\subsection{Interpretation and limitations}
				\label{app:faithfulness_limitations}
				
				The perturbation experiment should be interpreted as a faithfulness test rather than as a causal intervention. A larger error after masking highly attended information is evidence that the information associated with those attention weights is functionally relevant to the trained model under the chosen perturbation.
				
				This analysis has two limitations worth stating explicitly. First, attention weights can correlate with other properties of the input, most importantly recency: in the Text$\rightarrow$Numerical direction, the weeks receiving the largest attention weights are overwhelmingly the most recent weeks in the lookback window (Section 5.6). Because recent information is informative for most forecasting models regardless of attention, part of the large masking effect in this direction is confounded with simply removing recent data, rather than being attributable solely to the model having correctly identified informative weeks. We flag this explicitly as a recency confound: it limits how strongly the Text$\rightarrow$Numerical result can be read as evidence of attention doing something beyond tracking recency, even though the ordering Top-$k >$ Random-$k >$ Bottom-$k$ still holds by construction. Second, masking removes information entirely rather than perturbing it smoothly, so the resulting change in MSE reflects both the value of the missing information and the effect of an out-of-distribution input the model was not trained on. With these caveats in mind, we describe the Text$\rightarrow$Numerical pathway as the more functionally informative of the two under this perturbation protocol, while noting that cleanly separating attention-driven relevance from recency would require a masking design -- and likely a larger, more temporally diverse dataset -- that varies recency and attention independently.
				
				\subsection{Integrated Gradients}
				
				As a complementary feature-level analysis, Integrated Gradients was used to attribute the forecast to individual numerical and text-derived inputs. This analysis differs from CMA perturbation in its unit of attribution: rather than ranking complete historical weeks according to cross-modal attention, it assigns attribution scores to individual input features. The resulting global attribution visualization is provided in the main manuscript. We use it as complementary evidence rather than as a replacement for the targeted CMA perturbation analysis.

				\section{Summary of Supplementary Evidence}
				\label{app:summary}
				
				The supplementary analyses collectively support four conclusions that reinforce and extend the main-text findings. \textbf{First}, the principal DSA result is stable rather than a product of one favorable initialization: DSA wins directionally in all ten seeds against each of the three core baselines, the final training configuration was fixed before this multi-seed comparison (Table~\ref{tab:training_config}), and the Newey--West-corrected DM statistics, effect sizes, and bootstrap rank probabilities (Table~\ref{tab:supp_dm}, Table~\ref{tab:effect_size}, Table~\ref{tab:rank_probability}) all point the same direction even where individual DM tests fall short of conventional significance. \textbf{Second}, DSA's architecture is deliberate rather than incidental: the causal mask, mean pooling, and the numerical-encoder depth were each selected by an ablation that showed the alternative to be worse (Table~\ref{tab:supp_architecture}). \textbf{Third}, the advantage is attributable to the fusion mechanism specifically, and is robust to how that mechanism is instantiated: DSA outperforms the strongest fusion baseline even with BioLinkBERT frozen (Figures~\ref{FIG:12}--\ref{FIG:14}), and remains competitive when BioLinkBERT is replaced by general-domain encoders of different sizes (Table~\ref{tab:supp_text_encoders}; Figure~\ref{FIG:15}), indicating that neither a specific language model nor its fine-tuning is doing the load-bearing work. A single-seed test of the attention mechanism's directionality (Table~\ref{tab:supp_directional_cma}) points the same way: both single-direction variants underperform full bidirectional DSA, though this comparison has not been repeated across seeds. \textbf{Fourth}, the CMA faithfulness experiment provides direct functional evidence, not just an accuracy proxy, that the learned attention is doing something real: masking the most-attended weeks degrades forecasts significantly more than masking random or least-attended weeks (Table~\ref{tab:supp_perturbation}), an effect we interpret conservatively as functional relevance rather than as a causal claim.
				
				Taken together, these supplementary analyses reinforce the central claim of the paper on firmer ground than the main-text results alone would provide, while also placing honest boundaries around it: DSA is well supported as a robust, reproducible multimodal forecasting architecture for the tested 12-week ILI forecasting problem, with evidence spanning reproducibility, ablation, and faithfulness testing rather than resting on a single aggregate metric. At the same time, we do not claim universal superiority across every epidemic phase, geography, forecast horizon, or dataset outside the tested benchmark — the phase-stratified and cross-geography results in particular (Table~\ref{tab:supp_phase}, Table~\ref{tab:supp_africa}) identify specific, named conditions under which this is not yet established, which we regard as a more useful and durable claim than an unqualified one.

				\section*{Declaration of Competing Interest}
				The authors declare that they have no known competing financial interests or personal relationships that could have appeared to
				influence the work reported in this paper.
				
				\section*{Data Availability}
				The Health\_US and Health\_Africa datasets used in this study are subsets of the publicly available Time-MMD multimodal time-series benchmark\cite{liu2024time}. Code implementing the DSA architecture and the experiments reported in this paper will be available publicly upon publication.
				
				\section*{Declaration of generative AI and AI-assisted technologies in the manuscript preparation process}
				During the preparation of this work, the authors used ChatGPT (OpenAI) to assist with language refinement, manuscript organization, and LaTeX editing. After using this tool, the authors reviewed and edited the material as needed, verified the scientific content and references, and take full responsibility for the content of the published article.
				
				\printcredits
				
				\bibliographystyle{model1-num-names}

				\bibliography{cas-refs}

@ARTICLE{shaman2012forecasting,
	title={Forecasting seasonal outbreaks of influenza},
	author={Shaman, Jeffrey and Karspeck, Alicia},
	journal={Proceedings of the National Academy of Sciences},
	volume={109},
	number={50},
	pages={20425--20430},
	year={2012},
	publisher={National Academy of Sciences}
}

@ARTICLE{brownstein2009digital,
	title={Digital disease detection—harnessing the Web for public health surveillance},
	author={Brownstein, John S and Freifeld, Clark C and Madoff, Lawrence C},
	journal={The New England journal of medicine},
	volume={360},
	number={21},
	pages={2153},
	year={2009}
}

@CONFERENCE{zeng2023transformers,
	title={Are transformers effective for time series forecasting?},
	author={Zeng, Ailing and Chen, Muxi and Zhang, Lei and Xu, Qiang},
	booktitle={Proceedings of the AAAI conference on artificial intelligence},
	volume={37},
	number={9},
	pages={11121--11128},
	year={2023}
}

@ARTICLE{nie2022time,
	title={A time series is worth 64 words: Long-term forecasting with transformers},
	author={Nie, Yuqi and Nguyen, Nam H and Sinthong, Phanwadee and Kalagnanam, Jayant},
	journal={arXiv preprint arXiv:2211.14730},
	year={2022}
}

@CONFERENCE{liu2024itransformer,
	title={itransformer: Inverted transformers are effective for time series forecasting},
	author={Liu, Yong and Hu, Tengge and Zhang, Haoran and Wu, Haixu and Wang, Shiyu and Ma, Lintao and Long, Mingsheng},
	booktitle={International conference on learning representations},
	volume={2024},
	pages={11116--11140},
	year={2024}
}

@ARTICLE{wu2021autoformer,
	title={Autoformer: Decomposition transformers with auto-correlation for long-term series forecasting},
	author={Wu, Haixu and Xu, Jiehui and Wang, Jianmin and Long, Mingsheng},
	journal={Advances in neural information processing systems},
	volume={34},
	pages={22419--22430},
	year={2021}
}

@CONFERENCE{zhou2022fedformer,
	title={Fedformer: Frequency enhanced decomposed transformer for long-term series forecasting},
	author={Zhou, Tian and Ma, Ziqing and Wen, Qingsong and Wang, Xue and Sun, Liang and Jin, Rong},
	booktitle={International conference on machine learning},
	pages={27268--27286},
	year={2022},
	organization={PMLR}
}

@ARTICLE{kitaev2020reformer,
	title={Reformer: The efficient transformer},
	author={Kitaev, Nikita and Kaiser, {\L}ukasz and Levskaya, Anselm},
	journal={arXiv preprint arXiv:2001.04451},
	year={2020}
}

@CONFERENCE{zhang2023crossformer,
	title={Crossformer: Transformer utilizing cross-dimension dependency for multivariate time series forecasting},
	author={Zhang, Yunhao and Yan, Junchi},
	booktitle={The eleventh international conference on learning representations},
	year={2023}
}

@ARTICLE{liu2022non,
	title={Non-stationary transformers: Exploring the stationarity in time series forecasting},
	author={Liu, Yong and Wu, Haixu and Wang, Jianmin and Long, Mingsheng},
	journal={Advances in neural information processing systems},
	volume={35},
	pages={9881--9893},
	year={2022}
}

@ARTICLE{wu2022timesnet,
	title={Timesnet: Temporal 2d-variation modeling for general time series analysis},
	author={Wu, Haixu and Hu, Tengge and Liu, Yong and Zhou, Hang and Wang, Jianmin and Long, Mingsheng},
	journal={arXiv preprint arXiv:2210.02186},
	year={2022}
}

@ARTICLE{he2023multi,
	title={Multi-step forecasting of multivariate time series using multi-attention collaborative network},
	author={He, Xiaoyu and Shi, Suixiang and Geng, Xiulin and Yu, Jie and Xu, Lingyu},
	journal={Expert Systems with Applications},
	volume={211},
	pages={118516},
	year={2023},
	publisher={Elsevier}
}

@ARTICLE{liu2021forecasting,
	title={Forecasting influenza epidemics in Hong Kong using Google search queries data: A new integrated approach},
	author={Liu, Yunhao and Feng, Gengzhong and Tsui, Kwok-Leung and Sun, Shaolong},
	journal={Expert Systems with Applications},
	volume={185},
	pages={115604},
	year={2021},
	publisher={Elsevier}
}

@CONFERENCE{li2026language,
	title={Language in the flow of time: Time-series-paired texts weaved into a unified temporal narrative},
	author={Li, Zihao and Lin, Xiao and Liu, Zhining and Zou, Jiaru and Wu, Ziwei and Zheng, Lecheng and Fu, Dongqi and Zhu, Yada and Hamann, Hendrik and Tong, Hanghang and others},
	booktitle={International Conference on Learning Representations},
	volume={2026},
	pages={24437--24484},
	year={2026}
}

@ARTICLE{lu2019vilbert,
	title={Vilbert: Pretraining task-agnostic visiolinguistic representations for vision-and-language tasks},
	author={Lu, Jiasen and Batra, Dhruv and Parikh, Devi and Lee, Stefan},
	journal={Advances in neural information processing systems},
	volume={32},
	year={2019}
}

@CONFERENCE{jia2024gpt4mts,
	title={Gpt4mts: Prompt-based large language model for multimodal time-series forecasting},
	author={Jia, Furong and Wang, Kevin and Zheng, Yixiang and Cao, Defu and Liu, Yan},
	booktitle={Proceedings of the AAAI conference on artificial intelligence},
	volume={38},
	number={21},
	pages={23343--23351},
	year={2024}
}

@ARTICLE{radford2019language,
	title={Language models are unsupervised multitask learners},
	author={Radford, Alec and Wu, Jeffrey and Child, Rewon and Luan, David and Amodei, Dario and Sutskever, Ilya and others},
	journal={OpenAI blog},
	volume={1},
	number={8},
	pages={9},
	year={2019}
}

@ARTICLE{kandula2019near,
	title={Near-term forecasts of influenza-like illness: An evaluation of autoregressive time series approaches},
	author={Kandula, Sasikiran and Shaman, Jeffrey},
	journal={Epidemics},
	volume={27},
	pages={41--51},
	year={2019},
	publisher={Elsevier}
}

@ARTICLE{vaswani2017attention,
	title={Attention is all you need},
	author={Vaswani, Ashish and Shazeer, Noam and Parmar, Niki and Uszkoreit, Jakob and Jones, Llion and Gomez, Aidan N and Kaiser, {\L}ukasz and Polosukhin, Illia},
	journal={Advances in neural information processing systems},
	volume={30},
	year={2017}
}

@CONFERENCE{yasunaga2022linkbert,
	title={Linkbert: Pretraining language models with document links},
	author={Yasunaga, Michihiro and Leskovec, Jure and Liang, Percy},
	booktitle={Proceedings of the 60th Annual Meeting of the Association for Computational Linguistics (Volume 1: Long Papers)},
	pages={8003--8016},
	year={2022}
}

@ARTICLE{liu2024time,
	title={Time-mmd: Multi-domain multimodal dataset for time series analysis},
	author={Liu, Haoxin and Xu, Shangqing and Zhao, Zhiyuan and Kong, Lingkai and Kamarthi, Harshavardhan and Sasanur, Aditya B and Sharma, Megha and Cui, Jiaming and Wen, Qingsong and Zhang, Chao and others},
	journal={Advances in Neural Information Processing Systems},
	volume={37},
	pages={77888--77933},
	year={2024}
}

@misc{openai2023chatgpt,
	author       = {{OpenAI}},
	title        = {Introducing ChatGPT},
	year         = {2023},
	howpublished = {\url{https://openai.com/index/chatgpt/}},
	note         = {Accessed: 2026-08-23}
}

@ARTICLE{loshchilov2017decoupled,
	title={Decoupled weight decay regularization},
	author={Loshchilov, Ilya and Hutter, Frank},
	journal={arXiv preprint arXiv:1711.05101},
	year={2017}
}

@ARTICLE{diebold2002comparing,
	title={Comparing predictive accuracy},
	author={Diebold, Francis X and Mariano, Robert S},
	journal={Journal of Business \& economic statistics},
	volume={20},
	number={1},
	pages={134--144},
	year={2002},
	publisher={Taylor \& Francis}
}

@ARTICLE{newey1986simple,
	title={A simple, positive semi-definite, heteroskedasticity and autocorrelationconsistent covariance matrix},
	author={Newey, Whitney K and West, Kenneth D},
	year={1986},
	publisher={National Bureau of Economic Research Cambridge, Mass., USA}
}

@book{cohen2013statistical,
	title={Statistical power analysis for the behavioral sciences},
	author={Cohen, Jacob},
	year={2013},
	publisher={routledge}
}

@ARTICLE{hansen2011model,
	title={The model confidence set},
	author={Hansen, Peter R and Lunde, Asger and Nason, James M},
	journal={Econometrica},
	volume={79},
	number={2},
	pages={453--497},
	year={2011},
	publisher={Wiley Online Library}
}

@ARTICLE{benjamini1995controlling,
	title={Controlling the false discovery rate: a practical and powerful approach to multiple testing},
	author={Benjamini, Yoav and Hochberg, Yosef},
	journal={Journal of the Royal statistical society: series B (Methodological)},
	volume={57},
	number={1},
	pages={289--300},
	year={1995},
	publisher={Wiley Online Library}
}

@ARTICLE{doms2018assessing,
	author  = {Doms, Colin and Kramer, Sarah C. and Shaman, Jeffrey},
	title   = {Assessing the Use of Influenza Forecasts and Epidemiological Modeling in Public Health Decision Making in the United States},
	journal = {Scientific Reports},
	volume  = {8},
	pages   = {12406},
	year    = {2018},
	doi     = {10.1038/s41598-018-30378-w}
}

@ARTICLE{kim2024predicting,
	title={Predicting seasonal influenza outbreaks with regime shift-informed dynamics for improved public health preparedness},
	author={Kim, Minhye and Kim, Yongkuk and Nah, Kyeongah},
	journal={Scientific reports},
	volume={14},
	number={1},
	pages={12698},
	year={2024},
	publisher={Nature Publishing Group UK London}
}

@CONFERENCE{kim2021reversible,
	title={Reversible instance normalization for accurate time-series forecasting against distribution shift},
	author={Kim, Taesung and Kim, Jinhee and Tae, Yunwon and Park, Cheonbok and Choi, Jang-Ho and Choo, Jaegul},
	booktitle={International conference on learning representations},
	year={2021}
}

@CONFERENCE{sundararajan2017axiomatic,
	title={Axiomatic attribution for deep networks},
	author={Sundararajan, Mukund and Taly, Ankur and Yan, Qiqi},
	booktitle={International conference on machine learning},
	pages={3319--3328},
	year={2017},
	organization={PMLR}
}

			\end{document}